\documentclass[12pt]{article}

\usepackage[a4paper,margin=25mm]{geometry}
\usepackage{url}

\usepackage[numbers,square,sort&compress]{natbib}

\usepackage[table,xcdraw]{xcolor}
\usepackage{graphicx}
\usepackage{multirow}
\usepackage{longtable}  % supplement Tables S3 and S5 (arydshln needs it loaded first)
\usepackage{textcomp}   % \textdegree in the supplement
\graphicspath{ {./figures/} }

\usepackage[
	colorlinks=true,
	linkcolor=blue,
	filecolor=blue,
	urlcolor=blue,
	citecolor=blue,
			]{hyperref}

\newcommand{\sref}[1]{\ref{#1}}
\newcommand{\methodname}{PGPI4MR}

\usepackage{xspace}
\newif\ifitaliclatin
\italiclatinfalse
\newcommand{\latinphrase}[1]{\ifitaliclatin\textit{#1}\else#1\fi}
\newcommand{\invivo}{\latinphrase{in~vivo}\xspace}

\usepackage{setspace}
\usepackage{booktabs}
\usepackage{arydshln}

\usepackage{pdflscape}

\usepackage{enumitem}
\usepackage[labelsep=period,labelfont=bf,justification=raggedright,singlelinecheck=false]{caption}

\newcommand{\tabnote}[1]{\par\smallskip\scriptsize #1}

\newcommand{\figcaption}[2]{%
	\refstepcounter{figure}%
	\label{#1}%
	\textbf{Fig.~\thefigure.} #2\par\medskip
}
\usepackage{amsmath}

\newcounter{suppmethod}
\newcounter{suppresult}
\makeatletter
\newcommand{\supp@heading}[3]{% #1 counter, #2 title, #3 label
	\stepcounter{#1}%
	\edef\supp@tmp{\noexpand\subsection*{\csname the#1\endcsname. \unexpanded{#2}}}\supp@tmp
	\protected@edef\@currentlabel{\csname the#1\endcsname}\label{#3}}
\newcommand{\suppmethod}[2]{\supp@heading{suppmethod}{#1}{#2}}
\newcommand{\suppresult}[2]{\supp@heading{suppresult}{#1}{#2}}
\newcommand{\suppsizes}{%
	\DeclareRobustCommand\normalsize{%
		\@setfontsize\normalsize\@xipt{13.6}%
		\abovedisplayskip 11\p@ \@plus3\p@ \@minus6\p@
		\abovedisplayshortskip \z@ \@plus3\p@
		\belowdisplayshortskip 6.5\p@ \@plus3.5\p@ \@minus3\p@
		\belowdisplayskip \abovedisplayskip
		\let\@listi\@listI}%
	\DeclareRobustCommand\small{%
		\@setfontsize\small\@xpt\@xiipt
		\abovedisplayskip 10\p@ \@plus2\p@ \@minus5\p@
		\abovedisplayshortskip \z@ \@plus3\p@
		\belowdisplayshortskip 6\p@ \@plus3\p@ \@minus3\p@
		\belowdisplayskip \abovedisplayskip}%
	\DeclareRobustCommand\footnotesize{%
		\@setfontsize\footnotesize\@ixpt{11}%
		\abovedisplayskip 8\p@ \@plus2\p@ \@minus4\p@
		\abovedisplayshortskip \z@ \@plus\p@
		\belowdisplayshortskip 4\p@ \@plus2\p@ \@minus2\p@
		\belowdisplayskip \abovedisplayskip}%
	\DeclareRobustCommand\scriptsize{\@setfontsize\scriptsize\@viiipt{9.5}}%
	\DeclareRobustCommand\tiny{\@setfontsize\tiny\@vipt\@viipt}%
	\DeclareRobustCommand\large{\@setfontsize\large\@xiipt{14}}%
	\DeclareRobustCommand\Large{\@setfontsize\Large\@xivpt{18}}%
	\DeclareRobustCommand\LARGE{\@setfontsize\LARGE\@xviipt{22}}%
	\setlength\parindent{17\p@}%
	\normalsize}
\makeatother

\begin{document}
	\begin{center}	
		\textbf{\Large {Prior-Guided Physics-Informed Deep Learning for Fast 4D MRI Reconstruction}} \\
		Mojtaba Safari$ ^1 $, Parag J. Parikh$ ^2 $, Shansong Wang$ ^1 $, Mingzhe Hu$ ^1 $, Philip Connell$ ^1 $, Stanley L. Liauw$ ^1 $, Ralph R. Weichselbaum$ ^1 $,  Zhen Tian$ ^1 $ and  Xiaofeng Yang$ ^{1, \ddagger} $ \\
    \end{center}
\begin{flushleft}
$ ^1 $Department of Radiation and Cellular Oncology, The University of Chicago, Chicago, IL, USA\\
$ ^2 $Department of Radiation Oncology, Henry Ford Health, Detroit, Michigan, USA\\

\vspace{1 cm}

$ ^\ddagger  $Corresponding Author: email: \url{xfyang@uchicago.edu} (XY)
\end{flushleft}

\newpage

\begin{abstract}
\noindent\textbf{Background and purpose:} Four-dimensional (4D) MRI captures physiological motion and is valuable for both diagnostic imaging and radiotherapy guidance. However, achieving high temporal resolution requires acquiring each volume from only a small fraction of the raw data (k-space), which degrades image quality. We propose \methodname{} (prior-guided, physics-informed 4D MRI reconstruction), a deep learning method that leverages information from other respiratory phases or a setup scan, the first 30 s of the same scan, to improve 4D MRI reconstruction without extra scan time.

\noindent\textbf{Materials and methods:} \methodname{} is an unrolled network that reconstructs 16 respiratory phases (levels of breathing depth) from 60, 30, or 15 s of data, and single 3D volumes, each from 0.29--0.49 s of data, using a setup-scan image adjusted to the breathing amplitude of each volume. The networks were trained once, on other volunteers, and were not retrained for new volunteers. We evaluated \methodname{} on measured k-space from three datasets: free-breathing MR-linac (1.5T) and abdominal (3T) scans, and thigh scans with cuff compression (1.5T). All methods, including compressed sensing and the published 4D network MODEST, received identical samples and were scored by how well they predicted measured data withheld from all of them.

\noindent\textbf{Results:} In vivo, \methodname{} had the lowest held-out error for 4D MRI at every scan time, and its 16 phases were ready a median of 7.5--13 s after the scan on the 2 graphics processing units (GPUs) tested. Coupling the 16 respiratory phases reduced the error at 60 s by 9.1\% (95\% CI, 8.1\%--10.2\%; primary end point) compared with reconstructing each phase alone. For sub-second MRI, the error at 0.49 s was 15.5\% (95\% CI, 12.6\%--18.3\%) lower than that of the best comparison method, classical reconstruction toward the same prior, and on the faster GPU, every 0.29 s window was acquired and reconstructed within 0.5 s. Without retraining, \methodname{} matched the best methods on the 3T abdominal scans but not on the thigh scans; after retraining on thigh data, it reduced the held-out error by 6.6\% (95\% CI, 3.8\%--9.3\%) relative to the best comparison method.

\noindent\textbf{Conclusion:} Prior information from other respiratory phases or a setup scan allowed PGPI4MR to reconstruct 4D and sub-second MRI more accurately than compressed sensing and a published 4D network, within seconds and without extra scan time. This could support 4D MRI for treatment planning and 3D motion monitoring during MR-guided radiotherapy.
\end{abstract}

{\footnotesize \textbf{\textit{Keywords:}} {MR-linac, 4D MRI, Deep learning, Image reconstruction, Respiratory motion, Compressed sensing, Real-time imaging}}
\newpage 
\newpage

\section*{Introduction}
Respiratory motion is a central problem in radiotherapy of thoracic and abdominal tumors~\cite{keall2006tg76}. In radiology, physiological motion also degrades magnetic resonance imaging (MRI) of the chest and abdomen, which is therefore often acquired during breath holds or with respiratory triggering~\cite{kustner2020cinenet,lee2024diffusion}. Four-dimensional (4D) MRI instead captures this motion and is valuable for both diagnostic imaging~\cite{feng2016xdgrasp,kustner2020cinenet,lee2024diffusion} and radiotherapy guidance. An MRI scanner combined with a linear accelerator (MR-linac) can image soft-tissue before and during treatment~\cite{raaymakers2017first,winkel2019unity}. Before the beam is switched on, a respiratory-resolved 4D MRI, with one 3-dimensional (3D) image per respiratory phase, shows how the anatomy moves with breathing and supports planning and plan adaptation~\cite{stemkens2018nuts,vandelindt2018selfsorting}. During delivery, tumor trailing needs a new image at least once per second~\cite{fast2019trailing}, and real-time volumetric guidance~\cite{keall2025dose} needs 3D motion estimates within 0.5 s, including the acquisition~\cite{huttinga2022realtime}. Clinical systems therefore image 2-dimensional (2D) cine planes~\cite{lombardo2024motion}, because a 3D volume cannot be fully sampled that fast.

% Both 4D MRI and sub-second 3D MRI start from strongly undersampled k-space. Here, each of 16 respiratory states is undersampled 16.4- to 20.3-fold even after 60 s, and a 0.49 s window is undersampled 58-fold (median) on a coarser 4-mm grid. Parallel imaging~\cite{pruessmann1999sense} recovers little of this, because the 8 receive coils act like 1.86-2.21 independent channels (Supplementary Methods, \sref{sec:sampling}). Compressed sensing~\cite{lustig2007sparse} of one window has very few samples to work with, and methods that reconstruct the motion states together~\cite{feng2016xdgrasp,sarma2014lowrank} or fit a motion model~\cite{huttinga2020motus,huttinga2021lowrank} need the whole scan or long computation.

% Both 4D MRI and sub-second 3D MRI start from strongly undersampled k-space. Here, each respiratory state remains heavily undersampled even after the whole scan, and each short window is undersampled several times more, even on a coarser grid (Supplementary Methods, \sref{sec:states}). Parallel imaging~\cite{pruessmann1999sense} recovers little of this, because the receive coils act like only a few independent channels (Supplementary Methods, \sref{sec:sampling}). Compressed sensing~\cite{lustig2007sparse} of one window has very few samples to work with, and methods that reconstruct the motion states together~\cite{feng2016xdgrasp,sarma2014lowrank} or fit a motion model~\cite{huttinga2020motus,huttinga2021lowrank} need the whole scan or long computation.

Both 4D MRI and sub-second 3D MRI reconstruct images from strongly undersampled k-space. In 4D MRI, each respiratory phase remains far from fully sampled even when data from the entire scan are used; in sub-second 3D MRI, each short window has an acceleration factor several times higher still, despite its coarser grid (Supplementary Methods, \sref{sec:states}). Parallel imaging~\cite{pruessmann1999sense} recovers little of this gap because the receive coils behave as only a few independent channels (Supplementary Methods, \sref{sec:sampling}). Compressed sensing~\cite{lustig2007sparse} applied to a single window has very few samples to work with, while methods that jointly reconstruct all motion states~\cite{feng2016xdgrasp,sarma2014lowrank} or fit a motion model~\cite{huttinga2020motus,huttinga2021lowrank} require the whole scan or long computation times.

% Deep learning and prior-based methods add information that the samples lack. 
Deep learning and prior-based methods fill this gap by adding information that the samples lack. Unrolled networks alternate a learned regularizer with a data-consistency step that keeps the image close to the measured samples~\cite{SAFARI2026108291, hammernik2018variational,aggarwal2019modl,sriram2020e2evarnet}, and spatiotemporal networks also share information across motion states~\cite{terpstra2023modest,kustner2020cinenet,lonning2024drim,wang2026deepmocor}. Prior-based methods deform an earlier image to match new data~\cite{harris2016volumetric,shao2022ksregnet,shao2025dreme}, match new data to stored motion states~\cite{wu2023mrsigma}, or build or adapt patient-specific priors~\cite{atalik2026ltgvn,liu2024sparse,pan2025surface,xu2026inr}. Two questions remain open. First, the true image of moving anatomy cannot be measured, and scoring against processed or model-based reconstructions can inflate a method's gain~\cite{shimron2022crimes}. Second, it is unclear how much of a network's gain comes from the learned regularizer and how much from prior information, such as other motion states or an earlier image of the same person.

% We developed \methodname{} (prior-guided, physics-informed 4D MRI reconstruction), one physics-informed unrolled network design for both tasks. For 4D MRI, \methodname{} couples the 16 respiratory states. For sub-second MRI, it receives an image from a setup scan at the start of the session, moved to the breathing amplitude of each window. We tested it on public \invivo k-space data acquired using an MR-linac~\cite{terpstra2026fivedim,data_mrlinac} and on two external \invivo data sets. To answer the two questions, we scored all methods on measured k-space that none of them had seen. All methods received the same samples, coil maps, and, where used, prior. This study makes four contributions:

We developed \methodname{} (prior-guided, physics-informed 4D MRI reconstruction), a single unrolled network design for both 4D and sub-second 3D MRI. For 4D MRI, \methodname{} couples the 16 respiratory phases. For sub-second 3D MRI, it learns an image from a setup scan at the start of the session, deformed to the breathing amplitude of each window. We tested it on public \invivo MR-linac k-space data~\cite{terpstra2026fivedim,data_mrlinac} and on two external \invivo data sets. To address the first question, we scored all methods on measured k-space samples that none of them had seen. To address the second, all methods received the same samples, coil maps, and, where used, prior. This study makes four contributions:

\begin{enumerate}
	\item 4D MRI of 16 respiratory phases from 15 to 60 s of scanning, and a new 3D image from each 0.29 to 0.49 s of data, with acquisition plus reconstruction under 0.5 s for the shortest of these windows.
	\item An evaluation that needs no true image, with an analysis of how it responds to a prior.
	\item Comparisons against compressed sensing, classical reconstruction toward the same prior, and a published 4D network, which show that \methodname{} has the lowest errors on the MR-linac data and that prior information drives this advantage.
	\item Tests on 2 external \invivo data sets, one with fully sampled references.
\end{enumerate}

\section*{Materials and methods}
\subsection*{Data}
We used 3 public \invivo data sets. The main data set holds 60 s free-breathing scans of 10 volunteers on a 1.5 T Elekta Unity MR-linac (Elekta AB, Stockholm, Sweden)~\cite{terpstra2026fivedim,data_mrlinac,winkel2019unity} with 8 receive coils and 2.0 mm isotropic voxels. Its sampling is ordered pseudo-radial (OPRA) Cartesian sampling~\cite{joshi2022opra}: phase encodes are acquired along short L-shaped leaflets that rotate by the golden angle, and the data are stored in frames of about 98 ms. The data set provides ESPIRiT coil maps~\cite{uecker2014espirit} and the motion-resolved reconstruction (5D CMR-MOTUS, here MOTUS), which fits a reference image and a low-rank motion model to all 60~s~\cite{terpstra2026fivedim,huttinga2020motus,huttinga2021lowrank}. MOTUS use all the acquired samples, including those we held out for scoring, so we used it only as a reference for anatomy and motion.

To test transfer to another scanner and other anatomy, we used free-breathing abdominal scans at 3~T from Stanford~\cite{lee2024diffusion,data_stanford}, reconstructed one shot (0.31-0.43 s) at a time. We also used 1.5~T thigh scans of 9 volunteers under a pneumatic cuff from the University Medical Center (UMC) Utrecht~\cite{vanriel2026dataset,data_umc}, reconstructed in windows of 4 shots (0.69 s), and their fully sampled reference scans give an image-domain reference. The networks were applied to both zero-shot and, on the UMC data, also after training.

\subsection*{Reconstruction network}
For one respiratory phase or window, the samples $y$ are modeled as
\begin{equation}
	y = \mathcal{A}x + n, \qquad \mathcal{A} = \mathcal{M}\mathcal{F}\mathcal{S},
\end{equation}
where $x$ is the image, $\mathcal{S}$ applies the coil maps, $\mathcal{F}$ is the 2D Fourier transform over the 2 phase-encoding axes, $\mathcal{M}$ selects the sampled points, and $n$ is noise. \methodname{} follows model-based deep learning~\cite{aggarwal2019modl} (Fig.~\ref{fig:overview}B). Starting from the zero-filled image $\mathcal{A}^{H}y$, it repeats 2 steps 5 times:
\begin{equation}
	z_k = x_k + \mathcal{R}(x_k), \qquad x_{k+1} = \underset{x}{\arg \min} \,\, \|\mathcal{A}x-y\|_2^2 + \mu_k\|x-z_k\|_2^2 .
\end{equation}
The first step applies a learned regularizer $\mathcal{R}$, a residual U-Net~\cite{ronneberger2015unet} with 3 levels and 32 base channels, shared by all iterations. Each of its blocks applies a 3~$\times$~3~$\times$~3 spatial convolution, group normalization, and a Gaussian error linear unit (GELU) activation. The second step restores consistency with the measured samples by 3 conjugate-gradient iterations~\cite{pruessmann2001cgsense}, with a learned weight $\mu_k$. Here, $z_k$ is the network's current estimate of the image, not a prior image. For 4D MRI, $x$ holds all 16 phases, and no prior image is used. Instead, each 3~$\times$~3~$\times$~3 spatial convolution is followed by a convolution of size 3 across the phases~\cite{kustner2020cinenet}, initialized to zero so that training starts from a purely spatial network, and the prior information comes from the other phases. Without these, the same network reconstructs each phase separately; we used this 3D network as an ablation. For sub-second MRI, the moved setup prior enters $\mathcal{R}$ as 2 extra input channels (its real and imaginary parts) and does not enter the data-consistency step (Supplementary Methods, \sref{sec:network}).

\subsection*{4D MRI}
The frames were sorted into 16 equal-size respiratory phases by the amplitude of a self-gating navigator from the k-space center (Fig.~\ref{fig:overview}C). Each phase is one level of breathing depth, from the most exhaled to the most inhaled, with inhalation and exhalation pooled. We reconstructed the phases from the full 60 s and from the first 30 and 15 s of each scan.

\subsection*{Sub-second 3D MRI with a setup prior}
The first 30 s of each scan served as the setup scan, and the last 30 s were cut into consecutive windows of 3, 4, or 5 frames (0.29, 0.39, or 0.49 s), each reconstructed as one 3D volume on a 4-mm grid (2 mm right-left). The setup prior is the density-compensated zero-filled image of all setup samples, a blurred average of the anatomy over the setup breathing. It is moved to each window by
  \begin{equation}
      u(\mathbf{r}) = (\alpha - \bar{\alpha})\,\varphi(\mathbf{r}),
  \end{equation}
where $\varphi$ is one motion field per volunteer, $\alpha$ is the window's breathing amplitude, and $\bar{\alpha}$ is the mean amplitude of the setup scan, about where the averaged prior sits. We built 2 fields from setup k-space only. The registration-based field registers conjugate-gradient sensitivity encoding (CG-SENSE) images of 4 breathing bins of the setup scan. The learned field is a small coordinate network fitted so that the moved prior predicts the samples of each 0.49 s setup window (Fig.~\ref{fig:overview}A). In later windows, a learned amplitude module estimates $\alpha$ from the misfit between the moved prior and the window's samples (Fig.~\ref{fig:overview}B). \methodname{} uses the learned field, and Table~\ref{tab:table2} compares both (Supplementary Methods, \sref{sec:prior}).

\subsection*{Training on physics-based simulations}
The true image of moving anatomy cannot be measured, so all MR-linac networks were trained on physics-based MRI simulations of the measured scans: for each training volunteer, the MOTUS image is moved with the MOTUS motion model, k-space is simulated with the same forward model, and noise resampled from the measured data is added. Sharing the forward model makes simulated results optimistic~\cite{guerquinkern2012phantoms}. Our 4D networks were cross-validated in 5 folds of 8 training and 2 test volunteers. They were trained on 16 simulated motion states at random respiratory and cardiac positions with Poisson-disk masks, across per-state accelerations of 9 to 74, for 2,000 steps with Adam~\cite{kingma2014adam} at a learning rate of $3\times10^{-4}$ with cosine decay. Training used slabs of 48 readout positions (96 mm), and all 4D methods were reconstructed and scored on the central slab of 48 positions. The loss was the sum of the relative $\ell_1$ and $\ell_2$ errors of the image. Our sub-second networks were trained on whole volumes from windows of the scanner's own trajectory and then fine-tuned, at a learning rate of $1\times10^{-4}$, with random deformations and drifts of the prior. Every volunteer was scored by a network that had not seen it. All training was done once, before testing. On one NVIDIA RTX PRO 6000 Blackwell graphics processing unit (GPU), it took 5.3 to 5.5 h per 4D network and about 45 min per sub-second network, of which the final fine-tuning took 12 min. No network was trained or fine-tuned on the volunteer it reconstructed.

\subsection*{Comparison methods}
For 4D MRI, we compared the zero-filled image, CG-SENSE~\cite{pruessmann2001cgsense}, $\ell_1$-wavelet compressed sensing in the Berkeley Advanced Reconstruction Toolbox (BART)~\cite{lustig2007sparse,bart_software}, which reconstructs each phase separately, the published 4D network MODEST~\cite{terpstra2023modest}, our 3D network, and \methodname{}. For sub-second MRI, we compared the zero-filled image, CG-SENSE, BART, our network without a prior, classical conjugate gradient (CG) toward the moved prior,
\begin{equation}
	\hat{x} = \underset{x}{\arg \min}  \,\, \|\mathcal{A}x-y\|_2^2 + \lambda\|x-p\|_2^2 ,
\end{equation}
where $p$ is the setup prior moved along the same field by an amplitude fitted to the window, and \methodname{} with the moved prior. CG toward the moved prior uses the same information as \methodname{}, so their difference measures what the network adds. Weights were chosen on volunteers outside the test fold.

MODEST was run from its authors' original code, with its architecture and the published optimizer settings, and trained on our physics-based simulations (Supplementary Methods, \sref{sec:comparators}). The published end-to-end variational network (E2E-VarNet)~\cite{sriram2020e2evarnet} was run in the same way for sub-second MRI. It was designed for a fully sampled k-space center, which our sampling lacks, and it did not transfer to these data, so it is reported in the Supplementary Results only (\sref{res:published}).

\subsection*{Evaluation}
Real data have no true image, so we scored agreement with measured k-space that no method had seen~\cite{yaman2020ssdu}. For each volunteer, 12\% of the acquired phase-encoding grid points were drawn once for the whole scan and held out; the k-space center was never held out (Fig.~\ref{fig:overview}D). Their samples were withheld from every method, including the priors and the field fits; only the provided coil maps used all samples. The held-out k-space error compares each reconstruction, projected onto these points, with the samples measured there, averaging relative $\ell_1$ and $\ell_2$ errors over bands of k-space radius so that the few high-energy samples near the center do not dominate (Supplementary Methods, \sref{sec:heldout}). It is 0 for agreement with the measured samples and 2 for a zero image. Because the 4-mm sub-second grid does not reach the outer k-space, its score averages 5 bands rather than 8, so scores are compared between methods within a task, not between the 2 tasks (Supplementary Methods, \sref{sec:heldout}). Where a true image exists (physics-based simulations and fully sampled UMC references), we also report the normalized root-mean-square error (NRMSE) of the image inside the body, which complements this k-space measure. Organ motion beyond the heart was assessed as the head-foot shift of the lung-liver boundary across the 16 phases, compared with MOTUS (Supplementary Methods, \sref{sec:motion}).

Compute time, from the samples to the image in GPU memory, was measured offline on 2 GPUs: an NVIDIA RTX 6000 Ada and an RTX PRO 6000 Blackwell. We counted the time to an image as acquisition plus reconstruction, from the window start~\cite{huttinga2022realtime}, and also from the window middle, the mean age of the data.

\subsection*{Statistical analysis}
Paired comparisons used the volunteer as the unit (Stanford: the scan). We report the percentage reduction of the error, $100\,[1-\exp(\overline{\log(A/B)})]$, with a 95\% confidence interval (CI) from the \textit{t} distribution and the \textit{P} value of a 2-sided paired \textit{t}-test on the log ratios. The 2 end points were the 4D-versus-3D reduction at 60 s (primary) and the gain of classical CG from moving the prior along the registration-based field (confirmatory), each tested at $\alpha$ = 0.05. All other comparisons were registered (Supplementary Methods, \sref{sec:stats}) or exploratory (Table~\sref{tab:S5}).

\section*{Results}
\subsection*{4D MRI}
After an acquisition of 15 to 60 s, \methodname{} reconstructed all 16 phases in a median of 13.0 s on the Ada GPU and 7.5 s on the Blackwell GPU (Table~\ref{tab:table1}). In vivo, it had the lowest held-out error at every scan time. The primary end point was met: at 60 s, coupling the 16 phases lowered the held-out error by 9.1\% (95\% CI, 8.1\%-10.2\%; \textit{P}~$<$~0.001), and the reduction was similar with 30 and 15 s (Table~\ref{tab:table2}). BART had the second-lowest error, and \methodname{}'s margin over it grew as the scan shortened: 2.5\% at 60 s (95\% CI, 0.1\%-4.9\%; \textit{P}~=~0.045), an interval that included zero in 2 sensitivity analyses (Table~\sref{tab:S2}), and 6.5\% and 5.9\% at 30 and 15 s (both \textit{P}~$<$~0.001). \methodname{}'s image NRMSE was also the lowest, 18.7\% to 38.4\% below that of BART (all \textit{P}~$<$~0.001; Table~\sref{tab:S1}).

In Fig.~\ref{fig:figure2}A (60 s), the zero-filled image is hazy and MODEST's is smeared, whereas CG-SENSE, BART, the 3D network, and \methodname{} show the heart, the aorta, and the liver dome, with the most noise in CG-SENSE. In the simulation of this volunteer (Fig.~\ref{fig:figure2}B), \methodname{}'s error map is the darkest at 61 s and matches MODEST's at 30 and 15 s. On its test dataset, MODEST's held-out errors were close to those of the zero-filled image or CG-SENSE and above \methodname{}'s in each of them at every scan time. In the same dataset, \methodname{}'s NRMSE was 23.8\% lower than MODEST's at 61 s and 14.0\% lower at 30 and 15 s, and MODEST's was lower than BART's at 30 and 15 s.

\methodname{} also reproduced the respiratory motion of the liver (Fig.~\ref{fig:figure2}C and D), where lung-liver boundary followed MOTUS at 60, 30, and 15 s (\textit{r} = 0.99, 0.99, and 0.94), whereas the 3D network lost this motion as the scan shortened (\textit{r} = 0.73, 0.38, and 0.11).

\subsection*{Sub-second 3D MRI within 0.5 s}
With 0.29 s windows, acquisition plus reconstruction by \methodname{} took less than 0.5 s in every window on the Blackwell GPU (median reconstruction, 135 ms; slowest window, 0.455 s), and counted from the window middle, every window of every length was ready within 0.5 s on both GPUs (Fig.~\ref{fig:figure3}D; Table~\sref{tab:S3}). BART needed a median of 0.58 to 0.68 s to reconstruct one window. \methodname{} also had the lowest held-out error at every window length (Table~\ref{tab:table1}). Because this score favors images built on the setup prior (Supplementary Results, \sref{res:response}), the strongest comparison is CG toward the moved prior, the second-best method. \methodname{} had a 15.5\% lower held-out error than this method at 0.49 s (95\% CI, 12.6\%-18.3\%; \textit{P}~$<$~0.001) and an 11.5\% lower error at 0.29 s (\textit{P}~$<$~0.001). Its image NRMSE was also lower, by 10.3\% at 0.49 s (95\% CI, 7.1\%-13.4\%; \textit{P}~$<$~0.001) and by 3.4\% at 0.29 s, an interval that included zero ($-$3.4\% to 9.7\%; \textit{P}~=~0.28).

In the \invivo windows of Fig.~\ref{fig:figure3}A and B, the zero-filled, CG-SENSE, and BART images are dominated by undersampling streaks, and in B the network without a prior is blurred. Only the methods given the moved prior show the heart; CG toward this prior keeps the prior's noise-like texture, whereas \methodname{}'s image is smoother. Against the true image in Fig.~\ref{fig:figure3}C, \methodname{} has the darkest error map (NRMSE, 0.21), followed by CG toward the moved prior (0.25). In the one volunteer whose scan reached the upper abdomen, \methodname{} and the moved prior showed the aorta and the pancreas more clearly than the methods without a prior (Fig.~\sref{fig:S1}).

\subsection*{External data sets}
Without retraining, \methodname{} matched BART and CG toward the same prior on the Stanford abdominal scans (mean held-out error 1.35 for all 3; Table~\ref{tab:table1}; Table~\sref{tab:S4}). In Fig.~\ref{fig:figure4}A, these 3 methods show the kidneys and the spine similarly, whereas the zero-filled image is blurred and CG-SENSE is noisy. On the UMC thigh data, zero-shot \methodname{} had a 31.5\% higher image NRMSE than CG toward the same prior (95\% CI, 22.9\%-40.7\%; \textit{P}~$<$~0.001). After training on UMC volunteers in 3 folds, \methodname{} had a lower error than CG toward the same prior in every volunteer: 6.6\% in held-out error (\textit{P}~$<$~0.001) and 8.1\% in shading-insensitive NRMSE (\textit{P}~=~0.002; NRMSE, 4.4\%; 95\% CI, $-$0.4\% to 9.0\%; \textit{P}~=~0.066; Fig.~\ref{fig:figure4}C; Table~\sref{tab:S4}). In Fig.~\ref{fig:figure4}B, BART blurred the femur and a muscle septum, and CG toward the moved prior blurred septa at rest (arrows). At peak inflation, CG toward the moved prior and zero-shot \methodname{} showed a horizontal stripe below the femur that neither the reference nor trained \methodname{} showed.

\section*{Ablation}
The ablation shows where \methodname{}'s advantage comes from (Table~\ref{tab:table2}). In 4D MRI, coupling the phases lowered the held-out error by 7.8\% to 9.9\%; without this coupling, the 3D network had a 7.3\% higher error than BART at 60 s (\textit{P}~$<$~0.001). In sub-second MRI, the setup prior lowered the network's held-out error by 31.8\% (\textit{P}~$<$~0.001); without it, the network had an 8.2\% higher error than BART at 0.49 s (95\% CI, 3.5\%-13.1\%; \textit{P}~=~0.003) and an 8.6\% higher NRMSE (95\% CI, 1.3\%-16.5\%; \textit{P}~=~0.025). Moving the prior helped classical CG a little: along the registration-based field, it lowered the error by 0.51\% (95\% CI, 0.10\%-0.92\%; \textit{P}~=~0.020), which met the second confirmatory end point. \methodname{} had nearly the same error along the learned and the registration-based field (Supplementary Results, \sref{res:contribution}).

\section*{Discussion}

\methodname{} reconstructed both respiratory-resolved 4D MRI and sub-second 3D MRI from \invivo{} MR-linac data with lower errors than compressed sensing, classical reconstruction toward the same prior, and a published 4D network. It kept pace with the acquisition and returned a new 3D volume every 0.29 s, more often than tumor trailing requires~\cite{fast2019trailing}. The ablation traces this advantage to prior information, the input that \methodname{} was designed to use. Without it, the same network design had higher errors than compressed sensing. With only about 2 effective coil channels, a single phase or window cannot fill the k-space that it never sampled, so the missing information must come from other measurements of nearly the same anatomy, which the other respiratory phases and the setup scan provide. Given the same moved prior, \methodname{} still had a lower error than classical CG (Table~\ref{tab:table2}), so the network adds more than the prior itself. Without retraining, it matched the best methods on 3T abdominal scans, and on thigh scans, once trained on that anatomy~\cite{knoll2019generalization}, it had the lowest held-out error.

For clinical use, \methodname{} adds computation but no scan time. Its 4D MRI is reconstructed from the free-breathing scan itself, and its setup prior is the first 30 s of the same scan. The networks are trained once, before use, on other people's scans, and are not retrained for a new patient. For each patient, the setup prior and the motion field are computed automatically after the setup scan, in 8 to 14 s and 81 to 137 s, without contouring or manual tuning. In return, a 16-phase 4D MRI for planning or plan adaptation is ready a median of 7.5 to 13 s after a 15 to 60 s scan, whereas MOTUS, which also resolves cardiac motion, needs 6 min after a 1 min scan on its authors' GPU~\cite{terpstra2026fivedim}. For motion monitoring during delivery, a new 3D image is available every 0.29 s, within 0.5 s of the start of its acquisition on the faster GPU, where clinical systems now image 2D cine planes~\cite{lombardo2024motion}. In this study, all reconstructions ran offline, and the thigh data needed retraining, which took about 1.4 h per network.

% Because held-out k-space rewards images built on a prior, we judged \methodname{} against CG toward the same prior, and checked the ranking with image errors against simulated true images and fully sampled references.

The 2 tasks use prior information in different ways. In 4D MRI, coupling shares information across the respiratory phases, the principle behind the joint reconstruction of motion states~\cite{feng2016xdgrasp,sarma2014lowrank}. Unlike these methods, \methodname{} learns the coupling and reconstructs all 16 phases in seconds. The coupling also preserved the motion. The lung-liver boundary of \methodname{} followed MOTUS even at 15 s, whereas the 3D network lost this motion as the scan shortened. In sub-second MRI, the moved prior supplies the anatomy at the breathing position of the window, and the network corrects it with the samples of that window, which lowered the held-out error by 19.6\% relative to the moved prior alone (Supplementary Results, \sref{res:contribution}). In contrast to prior-based methods that deform an earlier image to match new data~\cite{harris2016volumetric,shao2022ksregnet}, \methodname{} therefore reconstructs each window under the guidance of the moved prior. In 4D MRI, the network received no prior image. Images from earlier fractions could provide one, as longitudinal priors do for other rapid MRI~\cite{atalik2026ltgvn}, once they are registered to the anatomy of the day.

MODEST had higher errors than \methodname{} on real data (Supplementary Results, \sref{res:published}). It was trained at one sampling density only and, unlike \methodname{}, does not end with a data-consistency step. %These results, from 3 volunteers for MODEST, do not rank either network in its intended setting.

This study has 4 limitations. First, all subjects were healthy volunteers without tumors, and the MR-linac data came from one scanner and one sequence; patients may breathe less regularly. Second, the sub-second volumes depend on the moved setup prior, so motion beyond the range of the setup scan may be underestimated (Supplementary Results, \sref{res:contribution}). Third, the true images of the volunteers are unknown. The held-out error responds only weakly to misplaced anatomy (Supplementary Results, \sref{res:response}), and in vivo, geometric accuracy was assessed only as liver motion against the MOTUS model in one volunteer. Fourth, the MR-linac networks were trained on 8 volunteers, with settings tuned partly on volunteers who later served as test volunteers; without these volunteers, the main comparisons were similar (Table~\sref{tab:S2}). Patient studies with target-tracking error as the end point should follow.

\section*{Conclusion}

\methodname{} reconstructed respiratory-resolved 4D MRI and sub-second 3D MRI from MR-linac data with lower errors than compressed sensing, classical reconstruction toward the same prior, and a published 4D network. Prior information from the other respiratory phases or a setup scan drove this gain without extra scan time, which supports prior-guided reconstruction for 4D MRI in treatment planning and for 3D motion monitoring in MR-guided radiotherapy.

\section*{CRediT authorship contribution statement}
\noindent \textbf{Mojtaba Safari:} writing -- original draft, methodology, conceptualization, visualization, software, formal analysis, data curation. \textbf{Parag J. Parikh:} conceptualization, writing -- review \& editing. \textbf{Shansong Wang:} data curation, software, conceptualization, writing -- review \& editing. \textbf{Mingzhe Hu:} software, visualization, writing -- review \& editing. \textbf{Philip Connell:} writing -- review \& editing. \textbf{Stanley L. Liauw:} writing -- review \& editing. \textbf{Ralph R. Weichselbaum:} writing -- review \& editing. \textbf{Zhen Tian:} resources, conceptualization, writing -- review \& editing, investigation.  \textbf{Xiaofeng Yang:} writing -- review \& editing, visualization, supervision, methodology, funding acquisition, resources, conceptualization.

\section*{Declaration of competing interest}
\noindent The authors declare that they have no known competing financial interests or personal relationships that could have appeared to influence the work reported in this paper.

\section*{Funding}
\noindent This research is supported in part by the National Institutes of Health under Award Numbers 1R01HL183223, R01DE033512 and R01CA272991.

\section*{Data availability}
\noindent All 3 data sets analyzed are public: the MR-linac k-space data of the 10 healthy volunteers (Zenodo record 21278895)~\cite{data_mrlinac}, the Stanford free-breathing abdominal data (Zenodo record 10607734)~\cite{data_stanford}, and the UMC Utrecht dynamic thigh data (Zenodo record 21194933, version 1.2)~\cite{data_umc}. The code and trained networks will be released in a public repository on publication; until then, they are available from the corresponding author on request.

\section*{Ethics statement}
\noindent This study is a secondary analysis of 3 public, de-identified data sets; no new data were acquired, and the authors had no access to identifying information, so the privacy rights of the subjects were observed.

\bibliographystyle{greenjournal}
\bibliography{reference}

\clearpage
\begin{landscape}
	\begin{table}[p]
		\begin{singlespace}
			\footnotesize
			\caption{Results of every method by acquisition time, in vivo and in physics-based simulation. Values are mean $\pm$ standard deviation over 10 volunteers (external data: see headers); arrows mark the better direction; bold, best; underlined, second best.}
			\label{tab:table1}
			\setlength{\tabcolsep}{2.3pt}\renewcommand{\arraystretch}{0.92}
			\makebox[\linewidth][c]{\begin{tabular}{@{}p{1.65cm} l ccc ccc c@{}}
			\toprule
			 & & \multicolumn{3}{c}{In vivo: held-out k-space error $\downarrow$} & \multicolumn{3}{c}{Simulation: NRMSE $\downarrow$} & Time (s) $\downarrow$ \\
			\cmidrule(lr){3-5}\cmidrule(lr){6-8}\cmidrule(l){9-9}
			Task & Method & 60 s & 30 s & 15 s & 61 s & 30 s & 15 s & Ada / Blackwell \\
			\midrule
			\multirow{7}{=}{\raggedright 4D MRI, 16 phases} & Zero-filled & 1.85\,$\pm$\,0.02 & 1.91\,$\pm$\,0.02 & 1.95\,$\pm$\,0.01 & 0.772\,$\pm$\,0.01 & 0.868\,$\pm$\,0.02 & 0.918\,$\pm$\,0.02 & 0.13 / 0.07 \\
			 & CG-SENSE & 1.63\,$\pm$\,0.05 & 1.74\,$\pm$\,0.04 & 1.83\,$\pm$\,0.04 & 0.463\,$\pm$\,0.05 & 0.653\,$\pm$\,0.04 & 0.774\,$\pm$\,0.04 & 4.7 / 2.6 \\
			 & BART & \underline{1.33}\,$\pm$\,0.10 & \underline{1.49}\,$\pm$\,0.09 & \underline{1.64}\,$\pm$\,0.07 & \underline{0.231}\,$\pm$\,0.05 & 0.420\,$\pm$\,0.07 & 0.622\,$\pm$\,0.07 & 25.5 / 17.5 \\
			 & MODEST & 1.90\,$\pm$\,0.04$^{\dagger}$ & 1.89\,$\pm$\,0.04$^{\dagger}$ & 1.83\,$\pm$\,0.05$^{\dagger}$ & 0.257\,$\pm$\,0.04$^{\dagger}$ & 0.327\,$\pm$\,0.05$^{\dagger}$ & 0.516\,$\pm$\,0.08$^{\dagger}$ & 27.7 / 21.4$^{\S}$ \\
			\cdashline{2-9}
			 & \rule{0pt}{2.5ex}\methodname{} without coupling (3D) & 1.43\,$\pm$\,0.09 & 1.55\,$\pm$\,0.10 & 1.67\,$\pm$\,0.09 & 0.264\,$\pm$\,0.06 & \underline{0.408}\,$\pm$\,0.08 & \underline{0.588}\,$\pm$\,0.10 & 10.5 / 5.5 \\
			 & \methodname{} (4D, phases coupled) & \textbf{1.30\,$\pm$\,0.08}$^{*}$ & \textbf{1.39\,$\pm$\,0.08}$^{*}$ & \textbf{1.54\,$\pm$\,0.07}$^{*}$ & \textbf{0.189\,$\pm$\,0.04}$^{*}$ & \textbf{0.263\,$\pm$\,0.06}$^{*}$ & \textbf{0.397\,$\pm$\,0.09}$^{*}$ & 13.0 / 7.5 \\
			\midrule
			 & & \multicolumn{3}{c}{In vivo: held-out k-space error $\downarrow$} & \multicolumn{3}{c}{Simulation: NRMSE $\downarrow$} & Time (ms) $\downarrow$ \\
			\cmidrule(lr){3-5}\cmidrule(lr){6-8}\cmidrule(l){9-9}
			 & & 0.49 s & 0.39 s & 0.29 s & 0.49 s & 0.39 s & 0.29 s & Ada / Blackwell \\
			\midrule
			\multirow{7}{=}{\raggedright Sub-second 3D MRI} & Zero-filled & 1.95\,$\pm$\,0.02 & 1.96\,$\pm$\,0.01 & 1.97\,$\pm$\,0.01 & 0.885\,$\pm$\,0.03 & 0.904\,$\pm$\,0.02 & 0.923\,$\pm$\,0.02 & $<$1 / $<$1 \\
			 & CG-SENSE & 1.81\,$\pm$\,0.06 & 1.83\,$\pm$\,0.05 & 1.87\,$\pm$\,0.04 & 0.731\,$\pm$\,0.05 & 0.766\,$\pm$\,0.04 & 0.799\,$\pm$\,0.04 & 31 / 20 \\
			 & BART & 1.61\,$\pm$\,0.11 & 1.64\,$\pm$\,0.10 & 1.68\,$\pm$\,0.10 & 0.614\,$\pm$\,0.06 & 0.663\,$\pm$\,0.06 & 0.710\,$\pm$\,0.05 & 658 / 582 \\
			 & CG toward the moved prior & \underline{1.31}\,$\pm$\,0.07 & \underline{1.32}\,$\pm$\,0.06 & \underline{1.32}\,$\pm$\,0.06 & \underline{0.276}\,$\pm$\,0.04 & \underline{0.280}\,$\pm$\,0.03 & \underline{0.283}\,$\pm$\,0.03 & 81 / 48 \\
			\cdashline{2-9}
			 & \rule{0pt}{2.5ex}\methodname{} without prior & 1.74\,$\pm$\,0.14 & 1.76\,$\pm$\,0.13 & 1.80\,$\pm$\,0.11 & 0.668\,$\pm$\,0.07 & 0.697\,$\pm$\,0.07 & 0.731\,$\pm$\,0.07 & 145 / 109 \\
			 & \methodname{} (moved prior) & \textbf{1.11\,$\pm$\,0.08}$^{*}$ & \textbf{1.13\,$\pm$\,0.08}$^{*}$ & \textbf{1.17\,$\pm$\,0.08}$^{*}$ & \textbf{0.248\,$\pm$\,0.04}$^{*}$ & \textbf{0.255\,$\pm$\,0.04}$^{*}$ & \textbf{0.275\,$\pm$\,0.04} & 188 / 136 \\
			\midrule
			 & & \multicolumn{2}{c}{Stanford abdomen, 6 scans} & \multicolumn{5}{c}{UMC thigh, 9 volunteers} \\
			\cmidrule(lr){3-4}\cmidrule(l){5-9}
			 & & \multicolumn{2}{c}{Held-out error $\downarrow$, one shot (0.31-0.43 s)} & \multicolumn{2}{c}{Held-out error $\downarrow$, 0.69 s} & \multicolumn{3}{c}{Image NRMSE $\downarrow$, 0.69 s} \\
			\midrule
			\multirow{8}{=}{\raggedright External \invivo data, zero-shot} & Zero-filled & \multicolumn{2}{c}{1.93\,$\pm$\,0.02} & \multicolumn{2}{c}{1.87\,$\pm$\,0.02} & \multicolumn{3}{c}{0.503\,$\pm$\,0.06} \\
			 & CG-SENSE & \multicolumn{2}{c}{1.61\,$\pm$\,0.16} & \multicolumn{2}{c}{1.62\,$\pm$\,0.01} & \multicolumn{3}{c}{0.382\,$\pm$\,0.03} \\
			 & BART & \multicolumn{2}{c}{\textbf{1.35\,$\pm$\,0.14}} & \multicolumn{2}{c}{1.41\,$\pm$\,0.02} & \multicolumn{3}{c}{0.298\,$\pm$\,0.02} \\
			 & CG toward the moved prior & \multicolumn{2}{c}{\textbf{1.35\,$\pm$\,0.14}} & \multicolumn{2}{c}{\underline{1.15}\,$\pm$\,0.04} & \multicolumn{3}{c}{\underline{0.267}\,$\pm$\,0.02} \\
			\cdashline{2-9}
			 & \rule{0pt}{2.5ex}\methodname{} without prior & \multicolumn{2}{c}{1.80\,$\pm$\,0.17} & \multicolumn{2}{c}{1.80\,$\pm$\,0.03} & \multicolumn{3}{c}{0.517\,$\pm$\,0.02} \\
			 & \methodname{} (moved prior) & \multicolumn{2}{c}{\textbf{1.35\,$\pm$\,0.14}} & \multicolumn{2}{c}{1.46\,$\pm$\,0.17} & \multicolumn{3}{c}{0.352\,$\pm$\,0.04} \\
			 & \parbox[t]{3.5cm}{\raggedright \methodname{}, trained on UMC volunteers\strut} & \multicolumn{2}{c}{NA} & \multicolumn{2}{c}{\textbf{1.08\,$\pm$\,0.06}$^{*}$} & \multicolumn{3}{c}{\textbf{0.256\,$\pm$\,0.03}} \\
			\bottomrule
			\end{tabular}}
			\tabnote{MODEST and time were not ranked. $^{*}$\textit{P}~$<$~0.05 against the second best (Table~\sref{tab:S5}). $^{\dagger}$MODEST: 3 test volunteers. $^{\S}$Per slab of 48 readout positions. NA, trained for UMC data only.}
		\end{singlespace}
	\end{table}
\end{landscape}
\clearpage
\begin{landscape}
	\begin{table}[p]
		\begin{singlespace}
			\footnotesize
			\caption{Ablation: the effect of each design choice in vivo (MR-linac) and in physics-based simulation. Each value is the percentage by which the tested setting lowered the error of the setting it is compared with (negative, a higher error), with its 95\% confidence interval.}
			\label{tab:table2}
			\setlength{\tabcolsep}{2.6pt}
			\begin{tabular}{@{}>{\raggedright\arraybackslash}p{2.6cm} l l l l r c r c@{}}
			\toprule
			 & & & & & \multicolumn{2}{c}{In vivo: held-out error} & \multicolumn{2}{c}{Simulation: NRMSE} \\
			\cmidrule(lr){6-7}\cmidrule(l){8-9}
			Question & Applied to & Tested & Compared with & Analysis & \% lower & 95\% CI & \% lower & 95\% CI \\
			\midrule
			\multirow[t]{3}{=}{Coupling the phases (4D MRI)} & \methodname{}, 60 s$^{\dagger}$ & Phases coupled (4D) & Each phase alone (3D) & Primary & 9.1\rlap{$^{*}$} & 8.1 to 10.2 & 28.8\rlap{$^{*}$} & 24.9 to 32.5 \\
			 & \methodname{}, 30 s & Phases coupled (4D) & Each phase alone (3D) & Registered & 9.9\rlap{$^{*}$} & 8.9 to 10.8 & 35.9\rlap{$^{*}$} & 32.4 to 39.3 \\
			 & \methodname{}, 15 s & Phases coupled (4D) & Each phase alone (3D) & Registered & 7.8\rlap{$^{*}$} & 6.2 to 9.4 & 33.2\rlap{$^{*}$} & 29.6 to 36.7 \\
			\cdashline{1-9}
			\multirow[t]{2}{=}{No prior information} & \rule{0pt}{2.5ex}4D MRI, 60 s & 3D network & BART & Exploratory & $-$7.3\rlap{$^{*}$} & $-$9.2 to $-$5.4 & $-$14.2\rlap{$^{*}$} & $-$19.6 to $-$9.1 \\
			 & Sub-second MRI & \methodname{} without prior & BART & Exploratory & $-$8.2\rlap{$^{*}$} & $-$13.1 to $-$3.5 & $-$8.6\rlap{$^{*}$} & $-$16.5 to $-$1.3 \\
			\cdashline{1-9}
			Adding the prior & \rule{0pt}{2.5ex}\methodname{} & Unmoved prior & No prior & Exploratory & 31.8\rlap{$^{*}$} & 28.7 to 34.8 & 60.4\rlap{$^{*}$} & 56.9 to 63.7 \\
			\cdashline{1-9}
			\multirow[t]{6}{=}{Moving the prior} & \rule{0pt}{2.5ex}Prior image alone & Learned field & Unmoved & Registered & 2.39\rlap{$^{*}$} & 1.81 to 2.96 & 4.32\rlap{$^{*}$} & 2.52 to 6.09 \\
			 & Prior image alone & Learned field & Registration-based field & Registered & 1.82\rlap{$^{*}$} & 1.56 to 2.08 & 1.44\rlap{$^{*}$} & 0.86 to 2.01 \\
			 & CG toward the prior & Registration-based field & Unmoved & Confirmatory & 0.51\rlap{$^{*}$} & 0.10 to 0.92 & 2.66\rlap{$^{*}$} & 1.19 to 4.10 \\
			 & \methodname{} & Learned field & Unmoved prior & Registered & 3.06\rlap{$^{*}$} & 2.42 to 3.69 & 4.82\rlap{$^{*}$} & 3.02 to 6.58 \\
			 & \methodname{} & Learned field & Registration-based field & Registered & 1.11\rlap{$^{*}$} & 0.83 to 1.38 & 1.17\rlap{$^{*}$} & 0.34 to 2.00 \\
			 & \methodname{}, learned field & Warp on & Warp switched off & Registered & 1.44\rlap{$^{*}$} & 0.66 to 2.21 & 3.05\rlap{$^{*}$} & 0.51 to 5.53 \\
			\cdashline{1-9}
			Network versus CG & \rule{0pt}{2.5ex}Same moved prior & \methodname{} & CG toward the moved prior & Registered & 15.5\rlap{$^{*}$} & 12.6 to 18.3 & 10.3\rlap{$^{*}$} & 7.1 to 13.4 \\
			\bottomrule
			\end{tabular}
		\tabnote{$^{*}$\textit{P}~$<$~0.05 (2-sided paired t-test; exact values in Table~\sref{tab:S5}). $^{\dagger}$Simulation: 61 s. Sub-second rows use 0.49 s windows. Primary and Confirmatory, the 2 end points fixed in advance; Registered, fixed in a dated internal analysis plan before the test data were scored (Supplementary Methods, \sref{sec:stats}); Exploratory, not registered. The learned-field rows are k-space results, not evidence of better motion.}
		\end{singlespace}
	\end{table}
\end{landscape}
\clearpage

\section*{Figure captions}
\figcaption{fig:overview}{Overview of \methodname{} on \invivo MR-linac data. (A) Setup scan (first 30 s): the setup prior and a motion field fitted to the setup k-space (not claimed to be the true motion). (B) Treatment: in each window of 0.29 to 0.49 s, the prior is moved to the breathing amplitude estimated from the window's k-space, and the unrolled network returns one 3D volume. (C) 4D branch: 16 respiratory phases reconstructed together. (D) Evaluation on \invivo data: 12\% of the acquired grid points, never the k-space center, are held out for scoring.}

\figcaption{fig:figure2}{Respiratory-resolved 4D MRI of one volunteer, 16 respiratory phases from 60, 30, or 15 s. (A) In vivo at 60 s, most exhaled phase. Whole width and MOTUS (context): MOTUS, dimmed outside the 96 mm block that all methods reconstructed; dashed lines, the other slices. Each image is scaled to MOTUS by one factor per slice; MOTUS was fitted on all data and is shown for anatomy only. Contours: the provided segmentation, moved by the MOTUS motion model; the liver's border with the lung follows the MOTUS image. (B) Simulation of the same volunteer and phase: images and error maps against the true image (0-50\% of the gray maximum). (C) In vivo, head-foot profiles through the liver dome on the sagittal slice of A, from the most exhaled (left) to the most inhaled phase; dashed line, the dome of the moved liver contour. (D) Head-foot shift of the liver dome at 60, 30, and 15 s (top to bottom); r, correlation with MOTUS.}

\figcaption{fig:figure3}{Sub-second 3D MRI (coronal slices). (A) In vivo, the first 0.29, 0.39, and 0.49 s windows after the setup scan of one volunteer; contours as in Fig.~\ref{fig:figure2}, moved to each window. (B) In vivo, the first 0.49 s window of a second volunteer. In A and B, each image is scaled to MOTUS by one factor; MOTUS is shown for anatomy only. (C) Simulation of the volunteer of B: images and error maps against the true image (0-50\% of the gray maximum). (D) Acquisition window and median compute time per window.}

\figcaption{fig:figure4}{External \invivo data sets. (A) Free-breathing abdomen at 3 T (Stanford), the normal-breathing scan with the most navigator motion; networks applied zero-shot. Reference: compressed sensing of the other shots of the same breathing bin, not a ground truth. (B) Thigh with a pressure cuff at 1.5 T (UMC), one 0.69 s window at rest and one at peak inflation; error maps (0-50\% of the gray maximum) against a fully sampled reference from a separate scan. Arrows: amber and red, structures blurred by BART and by CG toward the moved prior; white, a stripe absent from the reference. (C) All 9 UMC volunteers: thin lines, volunteers (gray, CG better); thick line, mean (Table~\sref{tab:S4}).}
\clearpage

%TC:ignore
\clearpage
\begin{figure}[p]
	\centering
	\includegraphics[width=\textwidth,height=0.9\textheight,keepaspectratio, draft=false]{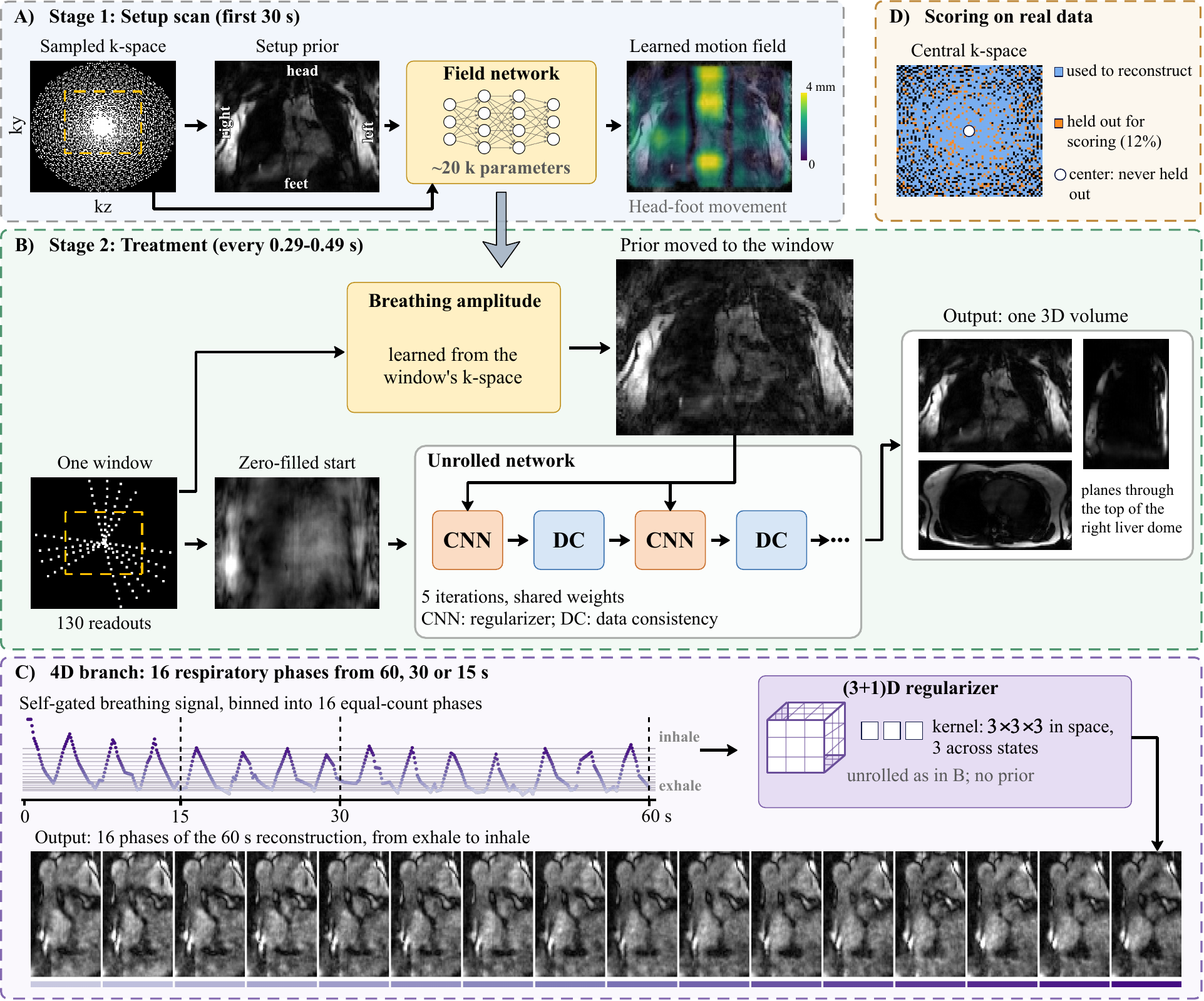}
	
	\medskip Fig. 1
\end{figure}
\clearpage
\begin{figure}[p]
	\centering
	\includegraphics[width=\textwidth,height=0.9\textheight,keepaspectratio, draft=false]{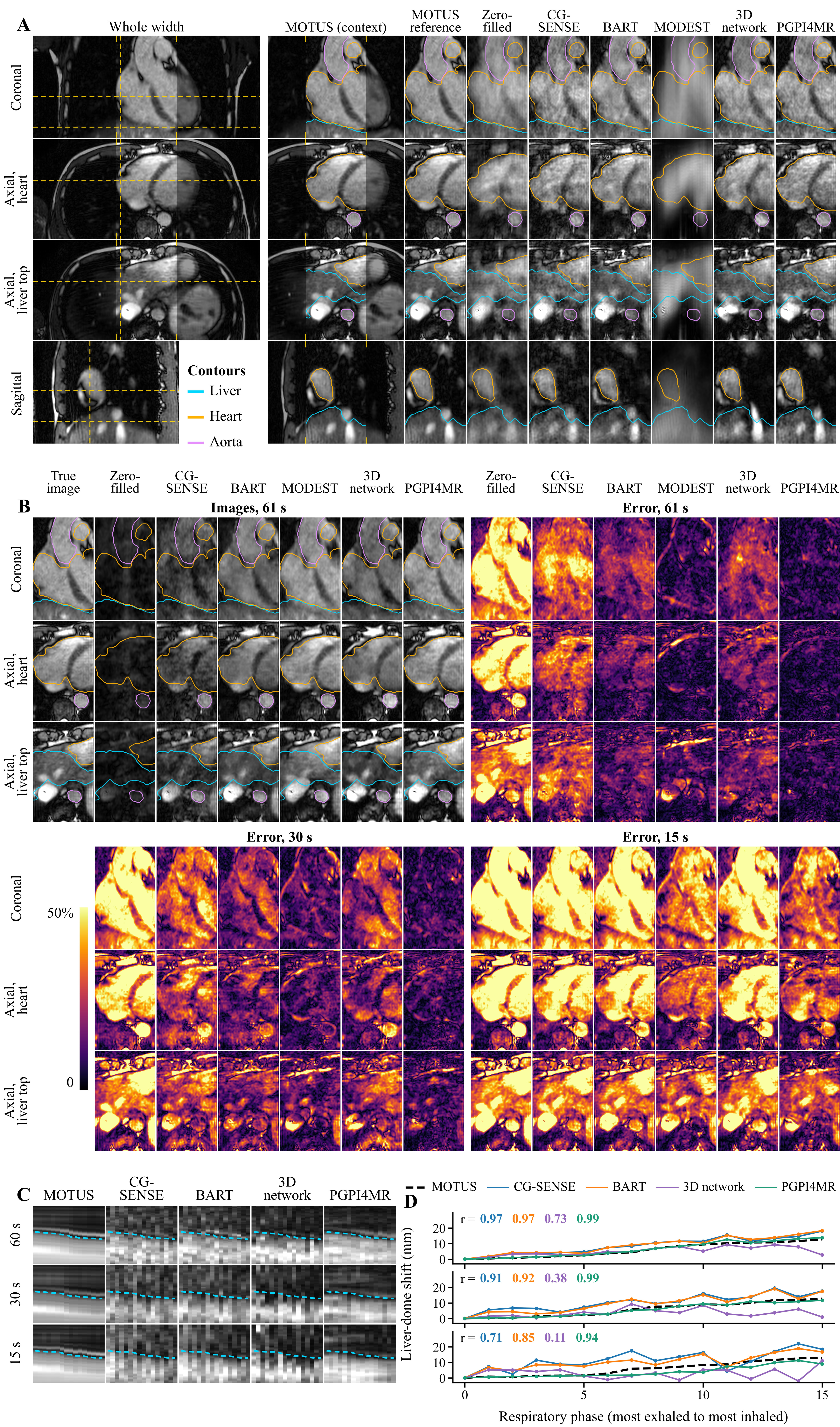}
	
	\medskip Fig. 2
\end{figure}
\clearpage
\begin{figure}[p]
	\centering
	\includegraphics[width=\textwidth,height=0.9\textheight,keepaspectratio, draft=false]{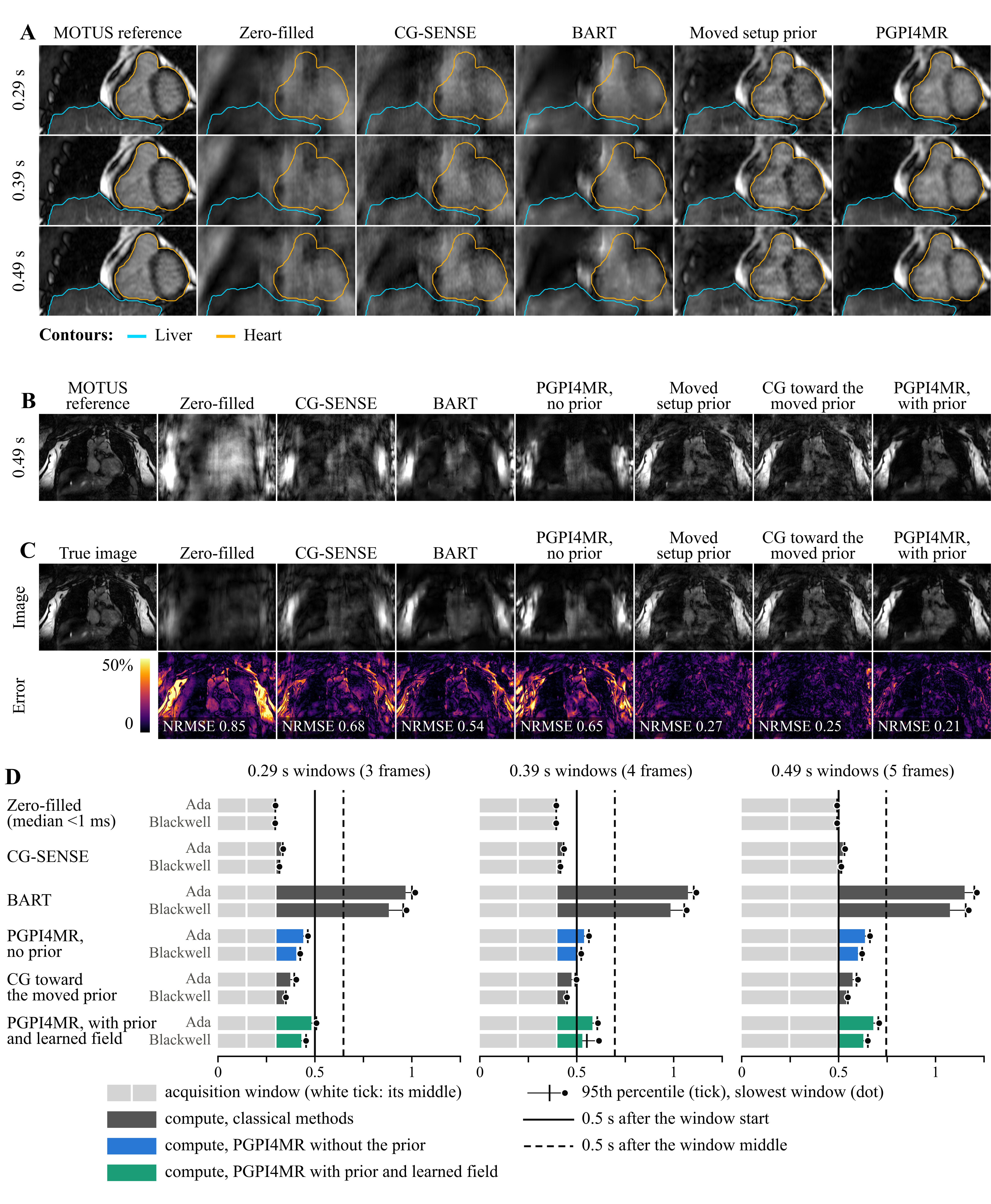}
	
	\medskip Fig. 3
\end{figure}
\clearpage
\begin{figure}[p]
	\centering
	\includegraphics[width=\textwidth,height=0.9\textheight,keepaspectratio, draft=false]{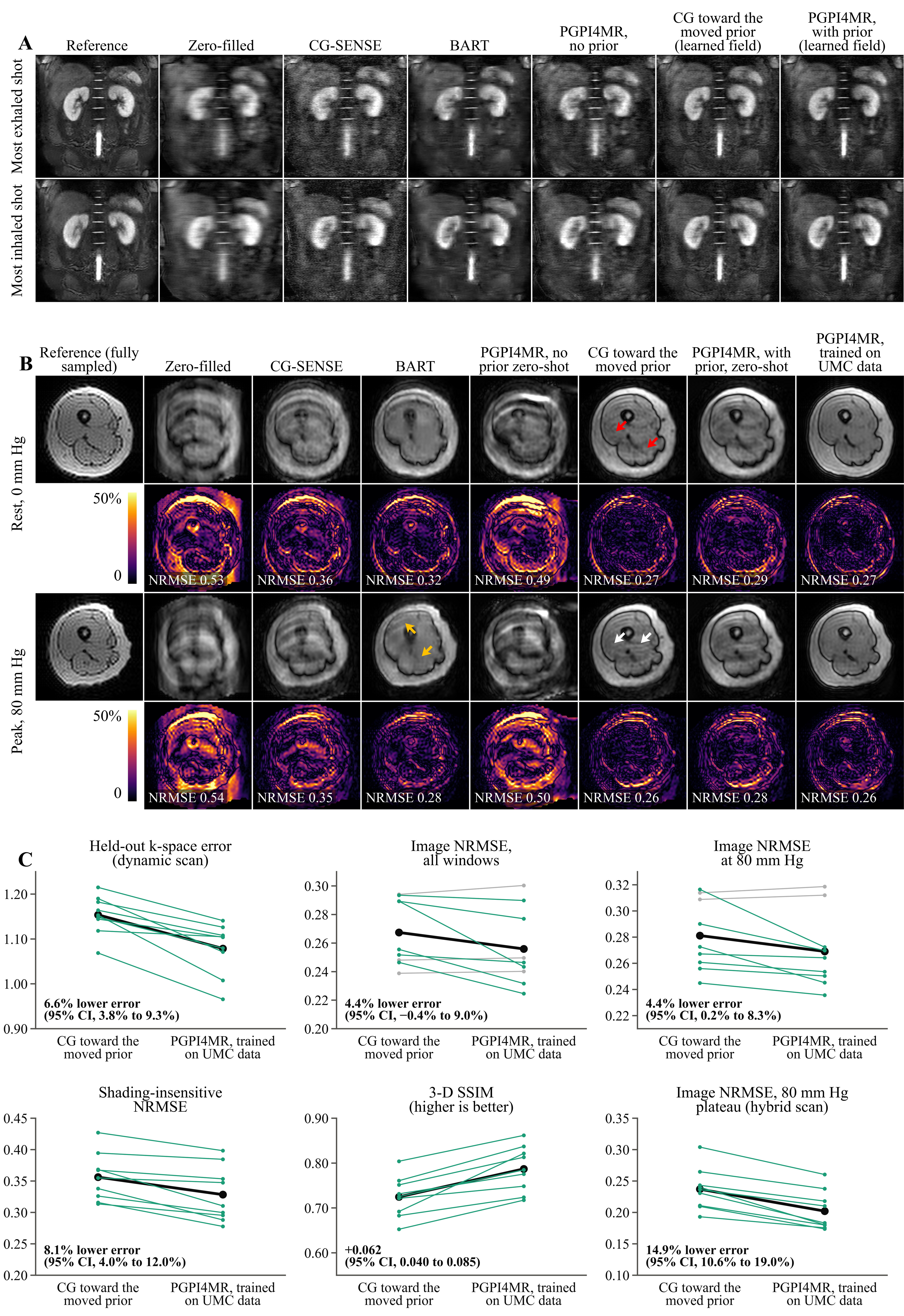}
	
	\medskip Fig. 4
\end{figure}
%TC:endignore

% =========================================================================================================
% Supplementary data: the body of PGPI4MR_supplement_v08.tex, typeset as in its own PDF (11 pt type,
% 20 mm margins). Its references are numbered with the main text's, in the list above.
% =========================================================================================================
\clearpage
\newgeometry{margin=20mm}
\suppsizes
\setcounter{table}{0}\renewcommand{\thetable}{S\arabic{table}}\renewcommand{\theHtable}{S\arabic{table}}
\setcounter{figure}{0}\renewcommand{\thefigure}{S\arabic{figure}}\renewcommand{\theHfigure}{S\arabic{figure}}
\captionsetup[table]{font=footnotesize}
\renewcommand{\tabnote}[1]{\par\smallskip\noindent\footnotesize #1}
\begin{center}
\textbf{\large Supplementary data}\\[4pt]
Prior-Guided Physics-Informed Deep Learning for Fast 4D MRI Reconstruction
\end{center}

\section*{Supplementary Methods}

\suppmethod{Sequence, sampling, and coil encoding}{sec:sampling}
The MR-linac scans used a balanced steady-state free precession sequence with a repetition time of 3.77-3.80 ms, an echo time of half the repetition time, and a flip angle of 50\textdegree. All acquired frames were used.

The effective number of independent receive channels was estimated by the participation ratio $1/\sum_i p_i^2$, where $p_i$ are the normalized eigenvalues of the coil-map Gram matrix inside the body. For the 8 receive coils, it was 1.86-2.21. No noise calibration scan was available, so the data were not prewhitened and the coil noise was assumed to be uncorrelated. Under this assumption, and for regular undersampling along the first phase-encoding axis, the median geometry factor (g-factor) was 1.05, 1.98, and 5.1 at 2-, 3-, and 4-fold undersampling. Coil encoding alone therefore supports about 2-fold undersampling, far less than the undersampling of a single respiratory phase or window.

\suppmethod{Respiratory phases and sub-second windows}{sec:states}
The self-gating navigator is the leading singular component of the readouts that pass through the k-space center, linearly interpolated to the frames without such a readout. In one volunteer, it agreed with the MOTUS respiratory signal at $|r| =$ 0.862. These readouts are too sparse to resolve cardiac motion, which is therefore averaged within each respiratory phase. The singular value decomposition that defines the navigator was computed from all 60 s, and the 30 and 15 s data were binned by the navigator values of their own frames.

The 16 respiratory phases are equal-count bins of the navigator amplitude; inhalation and exhalation are not separated. Each phase was trimmed at random to the sample count of the smallest phase, so that all phases share one acceleration. We define the per-phase acceleration as the number of grid points per readout position, including the corners outside the sampled ellipse, divided by the number of readouts per phase. It was 16.4-20.3, 31.1-48.0, and 65.6-85.5 at 60, 30, and 15 s before the hold-out, and 19-23, 36-55, and 77-101 after it. The 15 s data therefore lie beyond the range of accelerations used to train the 4D networks (9 to 74). Fewer phases would lower the acceleration in proportion; we used 16 phases to test the coupling under strong undersampling.

Each sub-second window was reconstructed on a grid of 4 mm (head-foot and anterior-posterior) by 2 mm (right-left). The median accelerations before the hold-out were 97, 73, and 58 for windows of 3, 4, and 5 frames.

\suppmethod{Held-out k-space error}{sec:heldout}
The trajectory visits the same phase-encoding grid points many times. If individual samples were held out, many of them would share their grid point with a sample used for the reconstruction and could be predicted by copying rather than by reconstruction. In one volunteer with 16 phases, 10\% of held-out samples would share their grid point with a reconstruction sample of the same phase (37\% in the innermost radius band), and 62\% of held-out grid points were acquired in at least one other phase. We therefore held out whole grid points, drawn once per volunteer over the whole scan, so that the held-out points are identical at 60, 30, and 15 s and in every window. The 4D hold-out (native grid) and the sub-second hold-out (4-mm grid) were drawn separately. The k-space center was never held out, because the image mean would then be unmeasured in every phase.

For a reconstruction $x$, the held-out k-space error is
\begin{equation*}
	E = \frac{1}{B}\sum_{b=1}^{B}\left(\frac{\|P_b(\mathcal{A}x-y)\|_1}{\|P_b y\|_1} + \frac{\|P_b(\mathcal{A}x-y)\|_2}{\|P_b y\|_2}\right),
\end{equation*}
where $P_b$ selects the held-out samples in radius band $b$. The band edges are 0.08, 0.19, 0.31, 0.43, 0.56, 0.70, and 0.84 of the native k-space radius (1 is the inscribed ellipse of the native grid), which gives 8 bands for 4D MRI. For sub-second MRI, the first 4 edges give 5 bands, because the k-space of the 4-mm grid extends only to about 0.5 along its 2 coarse axes. Band averaging keeps the few high-energy samples near the center from dominating the score. A single held-out grid point near the center carried 23\%-89\% of the held-out energy, and band averaging reduced the coefficient of variation of the 4D score across hold-out draws from 15\%-22\% to 1.7\%-3.9\%. The error is 0 for exact agreement with the measured held-out samples and 2 for a zero image. It evaluates the image only at acquired k-space locations, so it complements rather than replaces image-domain errors.

Because the measured samples contain noise, the error has a floor above 0. With the scanner's 4D sampling, the true image of the physics-based simulation, scored against its own k-space with Gaussian noise at the noise level of the simulation, had a held-out error of 0.52 and 0.73 in 2 volunteers, nearly the same at 30 and 15 s.

The scores of the 2 tasks are not comparable, because they average different bands. In the 4 inner bands, which have the same edges in both tasks, \methodname{} had a lower held-out error in 4D MRI at 60 s than in sub-second MRI at 0.49 s (0.86 versus 0.95), as did CG-SENSE (1.32 versus 1.76), in the same order as the simulated NRMSE (Table 1). Half of the 8 bands of the 4D score lie above 0.43 of the radius, where the mean error was 1.73 for \methodname{} and 1.95 for CG-SENSE, whereas only 1 of the 5 bands of the sub-second score does.

The provided MOTUS images equal the MOTUS reference image warped by its low-rank motion model, to a relative difference of $2\times10^{-7}$. They are therefore model outputs fitted to all 60 s, including the held-out samples, rather than measurements. In the 4D comparison, each MOTUS phase is the reference image warped to the mean respiratory and cardiac amplitudes of the frames in its bin, so it shows a single cardiac configuration, whereas every reconstruction averages over the cardiac cycle.

\suppmethod{Network and training}{sec:network}
The regularizer of \methodname{} (prior-guided, physics-informed 4D MRI reconstruction), the 3-level residual U-Net with 32 base channels described in the main text, has 6.49 million parameters in the 4D network and 5.84 million in the 3D network. In the 4D network, the convolution across the phases is zero-padded at the first and last phase and added to the output of each block. The current image enters the regularizer as 2 channels (its real and imaginary parts), and in the sub-second networks the moved prior adds 2 further channels. The data are scaled by the maximum of $|\mathcal{A}^{H}y|$, and the prior by its own maximum. Whole 4D volumes were reconstructed only to measure the compute times in Table~\ref{tab:S3}, and the networks processed them in slabs of 48 positions. The sub-second networks were trained and scored on the whole readout axis.

With Adam~\cite{kingma2014adam}, the learning rate of $\mu$ was $3\times10^{-3}$, the cosine decay of the 4D networks followed a 100-step warm-up, and gradients were clipped at 1.0. The weight $\mu$ was initialized at 0.05, and each training sample contained 16 simulated motion states of one volunteer in one slab. The sub-second networks have the same design without the convolution across phases, and their scores were averaged over 2 training runs.

The simulation shares the forward model and the coil maps with the reconstruction, and it omits off-resonance changes, eddy currents, motion within a window, and sliding motion. Its true image carries the noise of the MOTUS reference image. Its 4D test states combine 4 respiratory and 4 cardiac amplitudes and use Poisson-disk masks rather than the scanner's binned sampling. The masks undersample each state 18-, 37-, or 74-fold, the sample density of 16 respiratory phases from 61, 30, or 15 s of scanning, so simulated 4D results are labeled 61 s rather than 60 s. Its sub-second priors come from the measured setup scans, so the prior and the true image share anatomical features and noise, which favors methods that use the prior. As on real data, the sub-second simulated test set excludes each volunteer's held-out grid points from the windows and from the prior. Two stress tests added either a smooth random deformation on a 32-mm grid with slow drift, which a single motion mode cannot represent, or breathing amplitudes beyond the range of the setup scan.

\suppmethod{Setup prior, motion fields, and amplitude}{sec:prior}
The displacement model of the main text has a single motion mode. In the MOTUS motion model, the leading singular component carries 99.34\%-99.99\% of the energy of the respiratory component. A second mode, fitted in the same way, changed the error on a validation subset of setup samples withheld from the fit by a median factor of 0.995 on the MR-linac data and 0.997 on the Stanford data, that is, by less than 1\%. A single mode cannot represent hysteresis or drift.

The registration-based field registers CG-SENSE images of 4 navigator-sorted setup bins on the 4-mm grid to leave-one-bin-out templates and fits a least-squares slope against the setup navigator. The learned field is a coordinate network with about 20,000 parameters, evaluated on a 16-mm control grid. Fourier features of the coordinates with periods of 256, 128, and 64 mm (21 inputs) feed 2 hidden layers of 128 units with GELU activations and a linear layer with 3 outputs. The field is therefore smooth and does not represent sliding at the pleura. During the fit, each 0.49 s setup window had its own free amplitude, and each batch contained 8 setup windows. The field was fitted with Adam (learning rate, $1\times10^{-3}$ for the network and $1\times10^{-2}$ for the amplitudes) for 1,000, 500, and 500 steps at smoothness weights of 30, 3, and 0.3. This schedule was chosen on a validation subset of setup samples withheld from the fit, on which the fitted field lowered the error to 0.970-0.986 of that without a field. Because the fit minimizes the same k-space error that is used for scoring, an advantage of the learned field on this measure is expected; its image-domain behavior is therefore reported in simulation (Supplementary Results, R3). The fit took 81-137 s of GPU time per volunteer. The registration-based field needs no such fit, and \methodname{} had nearly the same error along it (Table 2).

Both fields express the amplitude in millimeters of the 95th-percentile head-foot displacement of the body. The amplitude module, fine-tuned along the registration-based field, is therefore used unchanged along the learned field. The module receives 4 features of the misfit between the moved prior and the window's samples at the current amplitude: the misfit itself, its relative slope and curvature, and the position of the amplitude within the setup range. It updates the amplitude 6 times with shared weights (2 hidden layers of 64 units), within the setup range widened by 25\% of its span on each side. It was pre-trained for 20,000 steps on simulated windows moved by other volunteers' MOTUS models, with the amplitude that best fits each window's k-space as the label, and then fine-tuned together with the network.

\suppmethod{Comparison methods and published networks}{sec:comparators}
 The zero-filled image and CG-SENSE need no training; CG-SENSE used 20 conjugate-gradient iterations with a Tikhonov weight of 0.001. BART~\cite{bart_software} used 100 iterations of $\ell_1$-wavelet regularization~\cite{lustig2007sparse} over the spatial axes, with cycle spinning, for each phase or window separately, and its weight was chosen from 7 values on volunteers outside the test fold. CG toward the moved prior used 30 iterations, starting from the prior multiplied by one fitted complex scale, and its weight was chosen from 11 values in the same way. 

E2E-VarNet~\cite{sriram2020e2evarnet} was run from its authors' code in its published configuration (about 29.9 million parameters) and trained with its published settings. The original study used 1D masks with a fully sampled k-space center. Our sampling has no such center, so its coil-map network used a calibration band of 6 to 7 phase-encoding columns. It was trained without a prior in the same 5 folds as the sub-second \methodname{} networks, on simulated windows of the scanner's trajectory, each with the 2D sampling mask of its acquisition.

MODEST~\cite{terpstra2023modest} was run from the code released with its paper, adapted to 8 coils and 16 phases. Because its registration network has not been released, we used its published variant without motion fields. It was trained with its published optimizer settings on physics-based simulations and evaluated on the 3 volunteers used neither for its training nor for its model selection. Its output has an arbitrary intensity scale, so each phase and readout slice of its output was scaled by a least-squares fit to the samples it received. A 4D \methodname{} network of the same design was trained on the same simulated samples, so that the 2 networks could also be compared with identical training data.

MODEST used the provided coil maps and was scored with the same held-out error as \methodname{}. E2E-VarNet estimates its own coil maps, so on real data its held-out error was computed from the k-space that it predicts for each coil at the held-out points. In simulation, its root-sum-of-squares magnitude image was compared with the magnitude of the true image (magnitude NRMSE). This comparison used a separately simulated test set of the same windows, in which the held-out grid points were not removed and the noise was drawn independently.

\suppmethod{External data sets}{sec:external}
The Stanford data contain 10 scans of healthy volunteers from 6 sessions. We pooled the 6 scans with normal breathing, one per session; all 10 scans are listed in Table~\ref{tab:S4}. Each scan has 47-87 shots without diffusion weighting, one every 2.4 s, because diffusion-weighted shots are acquired between them. These are therefore single-shot reconstructions rather than a continuous sub-second series. The setup period, the first half of these shots, spans about 84-103 s, about 3 times the MR-linac setup scan, which favors the prior. The 25-55 coils were compressed to 16. The coil maps and the coil compression were computed from all shots, and 77 of the 168-188 held-out grid points lie in the coil-calibration region.

The UMC data contain 9 volunteers (3 women and 6 men). The UMC networks were trained for 20,000 steps in 3 folds, each with 5 training volunteers and a sixth for validation, and tested on the remaining 3. Both prior-based methods received the setup prior moved along the registration-based field, because, in a registered test on 2 volunteers, the learned field raised the image error of the moved prior (Supplementary Methods, S9). E2E-VarNet was applied zero-shot with each of its 5 MR-linac networks (one per fold), and the results were averaged per scan or volunteer, as were those of the \methodname{} networks. The UMC references come from separate scans, so even a long CG-SENSE reconstruction of the rest period had an NRMSE of 0.18-0.25 against the 0 mm Hg reference. This floor compresses the NRMSE differences between methods. The image NRMSE was computed inside a box around the thigh after one best-fit scale. The shading-insensitive NRMSE first divides each magnitude image by its local mean over about 14 mm, and the 3D structural similarity index (SSIM) follows the data set authors~\cite{vanriel2026dataset}. The hybrid scan has one cuff cycle with a held 80 mm Hg peak; its reference is a CG-SENSE image of the other plateau shots.

\suppmethod{Motion measures in vivo and in simulation}{sec:motion}
For each output on simulated data, we found the breathing amplitude at which the true image matched it best, defined as the lowest NRMSE over a fine grid of amplitudes. The tracking slope regresses these amplitudes on the true amplitudes of the simulation across the 16 windows of a volunteer; it is 1 when the output follows the breathing and 0 when it is static. The position error converts the difference between the 2 amplitudes into millimeters through the motion model of the simulation; we report its 95th percentile over the body in the worst of the 16 windows. The field error compares a motion field with the true motion of the simulation; it is 0 for a perfect field and 1 for a field no better than no field.

The liver motion of Fig. 2C and D was measured in vivo, on the sagittal slice of Fig. 2A. The liver contour provided with the data was moved to each of the 16 phases by the MOTUS motion model. At the most exhaled phase, a band of $\pm$8 voxels ($\pm$16 mm) around the liver's upper surface was selected wherever lung lay directly above it. The band covered the 22 readout positions of the slab nearest the volunteer's right side (the right hemidiaphragm). In each reconstruction, the head-foot shift of a phase was the shift of its image that best matched the most exhaled phase inside this band (normalized cross-correlation, search range $\pm$12 voxels, parabolic sub-voxel interpolation). The same estimator was applied to MOTUS, which served as the motion reference. Agreement was measured by the correlation $r$ between a method's 16 shifts and those of MOTUS, the slope of a line fitted to them (the amplitude ratio), and their root-mean-square difference. Each head-foot profile in Fig. 2C averages 5 neighboring columns along the anterior-posterior axis through the apex of the dome.

\suppmethod{Statistical analysis}{sec:stats}
The 2 end points of the main text and the tests of the learned field were fixed in dated internal analysis plans (not a public registry) before the final test data were scored, but after earlier results had been seen. Network settings were chosen during development, in part on volunteers who later served as test volunteers in the cross-validation (2 volunteers for 4D MRI and 4 for sub-second MRI). Table~\ref{tab:S2} repeats the main comparisons without them. The primary end point was met (Table 2). The confirmatory test of moving the prior for classical CG used an $\alpha$ of 0.05 and would also have been met at a Bonferroni-adjusted $\alpha$ of 0.025 (97.5\% CI, 0.03\%-1.00\%). Without the 2 volunteers whose results for this test had been seen before it was registered, its interval included 0 (0.43\%; 95\% CI, $-$0.03\% to 0.89\%). The registered secondary analyses of 4D MRI were the reductions at 30 and 15 s, the comparison with CG-SENSE, and the equivalence of 4D MRI from half the scan time with 3D MRI from the full scan time. The reductions by radius band, the comparison with MOTUS, which was fitted to the held-out samples, the gap between training and test volunteers, and the decomposition of the variance, which would require repeated training runs per fold, are not reported. The learned field was tested by 3 Holm-adjusted tests on MR-linac data, all met, and by 2 on Stanford data, of which the first was met. The comparisons of \methodname{} along the learned field with CG toward the same moved prior and with \methodname{} given the unmoved prior or the registration-based field, and the field error in simulation, were registered as descriptive. Two earlier designs of the learned motion did not meet their registered criteria, and on 2 UMC volunteers the learned field did not meet its rule, so the registration-based field was used on the UMC data. Two steps deviated from the plans: the descriptive analyses of the shorter windows were performed before the planned simulation analyses, and the planned verification of the analysis code was not complete when the fields were fitted. Exact \textit{P} values are given in Table~\ref{tab:S5}; all other analyses are exploratory. MODEST, scored in 3 volunteers, is reported descriptively, without an interval or a \textit{P} value.

The confidence intervals are conditional on the trained networks (1 training run per fold for the 4D networks and 2 averaged runs for the sub-second networks) and on one hold-out draw. Because the training sets of the folds overlap, the intervals describe the training procedure applied to these volunteers. Scores at 60, 30, and 15 s share their held-out samples, as do the 3 window lengths, so they are not independent. The UMC intervals treat the 9 volunteers as independent, although they were scored by 3 trained networks.

\section*{Supplementary Results}

\suppresult{Per-volunteer results and sensitivity analyses}{res:pervolunteer}
Table~\ref{tab:S1} gives the results of every method in every volunteer. Table~\ref{tab:S2} repeats the main comparisons with the fold, rather than the volunteer, as the unit and without the volunteers used during development. The reduction from coupling the phases kept its sign and size in both analyses. The margin of \methodname{} over BART at 60 s was less robust, and its interval included zero in both analyses, whereas the margins at 30 and 15 s remained above zero. Coupling acted mainly through anatomy shared between the phases: shuffling their order raised the error by only 1.6\%. In simulation, the 3D network had a higher NRMSE than BART at 61 s, a similar NRMSE at 30 s, and a lower NRMSE at 15 s.

On real data, the held-out error of MODEST at 60 and 30 s was close to that of the zero-filled image in each of its 3 test volunteers. At 15 s, its error was below that of the zero-filled image in all 3 volunteers (mean, 1.83 versus 1.96) and close to that of CG-SENSE (1.85). It was also lower than its error at 60 s in 2 of the 3 volunteers.

\suppresult{Response of the held-out k-space error to motion and to a prior}{res:response}
The held-out error responds only weakly to the position of the anatomy. Placed at its fitted amplitude along the learned field, the setup prior had a 3.7\% lower held-out error (95\% CI, 2.1\%-5.3\%) than when it was misplaced by three-quarters of its setup breathing range. On the Stanford data, \methodname{} with the unmoved prior had a 3.2\% lower held-out error with the prior in place than with the prior shifted 3 mm head-foot (95\% CI, 1.5\%-4.8\%). Differences of 1\% to 3\% between motion fields may therefore correspond to misplacements of a few millimeters.

The UMC data, which have fully sampled references, show how the held-out error treats prior-based methods. Fig.~\ref{fig:S2} plots the mean held-out error of each method against its mean image NRMSE. Without a prior on either side, both measures placed 2 methods close together: \methodname{} without a prior, trained on UMC volunteers, had a 0.2\% lower held-out error and a 1.5\% higher NRMSE than BART. The methods given the setup prior lay below the methods without a prior, that is, they had a lower held-out error at a similar image NRMSE. The moved setup prior alone, without reconstruction, had nearly the NRMSE of BART (0.296 versus 0.298) but a 10\% lower held-out error (1.26 versus 1.41). The unmoved setup prior alone had a lower held-out error than BART (1.28 versus 1.41) despite a higher NRMSE (0.345 versus 0.298), and zero-shot \methodname{} with the moved prior had a held-out error close to that of BART but a 17.7\% higher NRMSE. For this reason, prior-based methods were compared with CG toward the same moved prior, and pairs with and without a prior were judged by image errors where these were available.

\suppresult{Contribution of the window samples and of the motion fields}{res:contribution}
On real data at 0.49 s, \methodname{} with the unmoved prior had a 16.2\% lower held-out error than that prior alone (95\% CI, 14.0\%-18.3\%). Along the learned field, \methodname{} had a 19.6\% lower error than its moved prior alone (95\% CI, 16.6\%-22.6\%). The window samples therefore changed the image beyond the prior. In simulation, \methodname{} along the learned field had a 19.2\% lower NRMSE than its moved prior alone, even though the prior was scaled to the true image (Table~\ref{tab:S1}).

Moving the prior along the learned field lowered the held-out error of \methodname{} by 3.06\% (Table 2), and in a registered test, the same trained \methodname{} network had a 1.44\% lower error with its warp on than with its warp switched off, that is, than when it was given the unmoved prior at test time (Table 2). The learned field brought the setup prior closer to held-out k-space than the registration-based field, but \methodname{} had nearly the same error along both fields. These are k-space results and do not indicate more accurate motion. In simulation, the learned field was not closer to the true motion than the registration-based field (median field error, 0.853 versus 0.832, where 1 indicates no better than no field). In 2 UMC volunteers, it raised the image error of the moved prior by 2.3\% and 4.0\%.

In simulation, the output of \methodname{} followed the simulated breathing mainly through the moved prior, so its sub-second volumes reproduced the motion pattern of the setup scan, scaled for each window. At normal breathing, the median tracking slope of \methodname{} was 0.435 with the unmoved prior, 0.984 along the registration-based field, and 0.954 along the learned field. With the random deformation plus drift (Supplementary Methods, S4), the slopes were 0.314, 1.024, and 0.934, and with breathing beyond the setup range, 0.288, 0.780, and 0.723. Along the learned field, the position error at 0.49 s rose from 0.54 mm at normal breathing to 2.07 mm beyond the setup range (Table~\ref{tab:S1}). With the random deformation plus drift, the NRMSE of \methodname{} rose from 0.248 to 0.428, close to that of \methodname{} with the unmoved prior (0.431).

On real data along the learned field, the amplitude fitted to each window's k-space followed the self-gating navigator with a median $|r|$ of 0.931 and the MOTUS respiratory signal with a median $|r|$ of 0.952. Because the navigator is derived from the same k-space center samples, $r$ reflects timing rather than size. The amplitude estimated by the amplitude module matched this fit (median slope, 1.049; $r$ = 0.997). In the quarter of the body that moved most in MOTUS, however, the learned field moved the anatomy 0.627 times as far as the MOTUS field for the same navigator change.

\suppresult{Compute time and latency}{res:latency}
Latency, the time to an image (acquisition plus reconstruction), was counted from the window start~\cite{huttinga2022realtime} and from the window middle. Table~\ref{tab:S3} gives the compute time and latency of every method, window length, and GPU. Per 0.49 s window, \methodname{} with the moved prior needed a median of 188 ms on the Ada GPU and 136 ms on the Blackwell GPU. Counted from the window middle, its slowest image was ready after 0.465 s. Counted from the window start, its median latency was 0.43 to 0.68 s, and with 0.29 s windows on the Blackwell GPU every image was ready within 0.5 s (slowest window, 0.455 s). Stanford shots of 0.31 to 0.32 s were also ready within 0.5 s on both GPUs (slowest, 0.465 s); these shots are acquired only once every 2.4 s. Shots of 0.43 s were ready within 0.5 s of the shot middle (slowest, 0.358 s on the Ada GPU and 0.301 s on the Blackwell GPU) but not of the shot start (0.574 and 0.517 s). At 0.39 s on the Blackwell GPU, the slowest latency came from a single window that needed 222 ms of compute (95th percentile, 159 ms). Of the compute time of BART, 423-438 ms per window was spent in the solver at 100 iterations, an iteration count not tuned for speed. \methodname{} reconstructed a 16-phase 4D MRI in a median of 13.0 s on the Ada GPU and 7.5 s on the Blackwell GPU (ranges across volunteers, 10.6-17.5 s and 6.0-10.1 s, depending on the readout length). MODEST needed 27.7 s on the Ada GPU and 21.4 s on the Blackwell GPU per slab of 48 readout positions. A whole volume of 4 or 5 slabs would therefore take about 86 to 107 s on the Blackwell GPU.

\suppresult{External data sets}{res:external}
Table~\ref{tab:S4} lists the Stanford results of every scan and all UMC comparisons. On the Stanford data, \methodname{} without a prior had a 32.8\% higher held-out error than BART, as on the MR-linac data. Along the registration-based field, the held-out error of zero-shot \methodname{} differed from that of BART by 0.1\% (95\% CI, $-$3.6\% to 3.6\%). In the 2 deep-breathing scans, the amplitude along the registration-based field correlated with the navigator at $r$ = 0.834 and 0.478, compared with 0.50 and $-$0.27 along the learned field. In the registered test of the \methodname{} warp (on versus switched off), all 6 normal-breathing scans favored the warp, but one scan dominated the mean and the test was not significant (Holm-adjusted \textit{P}~=~0.18). Applied zero-shot, E2E-VarNet had the highest held-out error of all methods in each of the 6 Stanford scans and each of the 9 UMC volunteers, and the highest image NRMSE in each UMC volunteer. Without a prior on either side, zero-shot \methodname{} had a 29.9\% lower held-out error than E2E-VarNet on the Stanford data (95\% CI, 19.2\%-39.2\%). On the UMC data, zero-shot \methodname{} with the moved prior had a 58.2\% lower image NRMSE than E2E-VarNet (95\% CI, 54.1\%-61.8\%).

\suppresult{Published networks}{res:published}
E2E-VarNet is not shown in the main text. Its mean held-out error on real data was 3.31, 3.28, and 3.19 at 0.49, 0.39, and 0.29 s, and its mean magnitude NRMSE in simulation was 0.941, 0.931, and 0.918, on a separately simulated test set of the same windows without removal of the held-out grid points (Supplementary Methods, S6). Its compute time is given in Table~\ref{tab:S3} and its results on the external data sets in Table~\ref{tab:S4}. Without a prior on either side, \methodname{} had a 46.8\% lower held-out error than E2E-VarNet at 0.49 s on real data (95\% CI, 40.5\%-52.4\%) and a 41.7\% lower magnitude NRMSE in simulation (95\% CI, 37.4\%-45.7\%). A comparison of E2E-VarNet with \methodname{} given the moved prior was not made by held-out error, because this score favors images built on a prior (R2). In 4D MRI on real data, \methodname{} trained on the same simulations as MODEST had a lower held-out error than MODEST in each of the 3 volunteers at every scan time. The mean reductions were 33.9\%, 26.0\%, and 12.8\% at 60, 30, and 15 s.

In 8 windows of 2 test volunteers, the k-space predicted by E2E-VarNet had an error of 3.50 even on the samples it had received, compared with 0.20 for CG-SENSE and 1.17 for the zero-filled image. The phase of its image changed by a median of 168\textdegree{} between neighboring phase-encoding columns, close to a sign flip, and a median of 78\% of the energy of its complex image lay in the outermost of these columns. The data-consistency weight of its last cascade was $-$0.004, so its final step made little use of the data. Its magnitude image was also inaccurate: combined with the phase of CG-SENSE, it had a held-out error of 2.18, compared with 1.97 for the zero-filled image and 1.86 for CG-SENSE. In simulation, its magnitude NRMSE was 0.19 to 0.32 in 8 windows of 4 of its training volunteers but 0.91 to 0.96 in 4 windows of 2 test volunteers, higher than that of its zero-filled input (0.79 to 0.81). Across the 5 folds, its magnitude NRMSE in test volunteers was 0.83 to 1.14 (Table~\ref{tab:S1} gives the values at 0.49 s). On real data, it had a higher held-out error than the zero-filled image, CG-SENSE, and \methodname{} without a prior in all 30 combinations of volunteer and window length.

MODEST was also tested on simulated k-space with the scanner's ordered pseudo-radial (OPRA) sampling, in 2 of its 3 test volunteers. In one of them at 60 s, its held-out error was 1.931 in this simulation and 1.935 on real data, so the simulation matched its held-out error on real data. In this simulation, its complex NRMSE increased with more data, from 0.57 and 0.66 at 15 s to 0.75 and 0.81 at 60 s in the 2 volunteers. With Poisson-disk masks of the same sample counts, its NRMSE decreased with more data, from 0.67 at 15 s to 0.31 at 60 s in the first of the 2 volunteers. Its density compensation changed its held-out error on real data by at most 0.019. The 4D \methodname{} networks were also trained on Poisson-disk masks and transferred to real data (Table 1); unlike MODEST, they end with a data-consistency step.

\suppresult{Liver motion and organs below the heart}{res:organs}
In the volunteer of Fig. 2C, the lung-liver boundary of MOTUS moved 12.6 to 12.8 mm between the most exhaled and the most inhaled phase (Fig. 2C and D). \methodname{} followed this motion at $r$ = 0.99, 0.99, and 0.94 at 60, 30, and 15 s (amplitude ratio, 1.09, 0.94, and 0.75; root-mean-square difference, 0.8, 0.8, and 2.5 mm). The 3D network lost it as the scan shortened ($r$ = 0.73, 0.38, and 0.11; 3.2, 4.9, and 6.0 mm). CG-SENSE ($r$ = 0.97, 0.91, and 0.71; 3.1, 4.9, and 6.9 mm) and BART ($r$ = 0.97, 0.92, and 0.85; 3.3, 4.3, and 5.2 mm) followed the timing of the motion but with larger errors in each phase. These values come from one volunteer and one slice.

\begin{figure}[p]
\centering
\includegraphics[width=\textwidth,height=0.82\textheight,keepaspectratio]{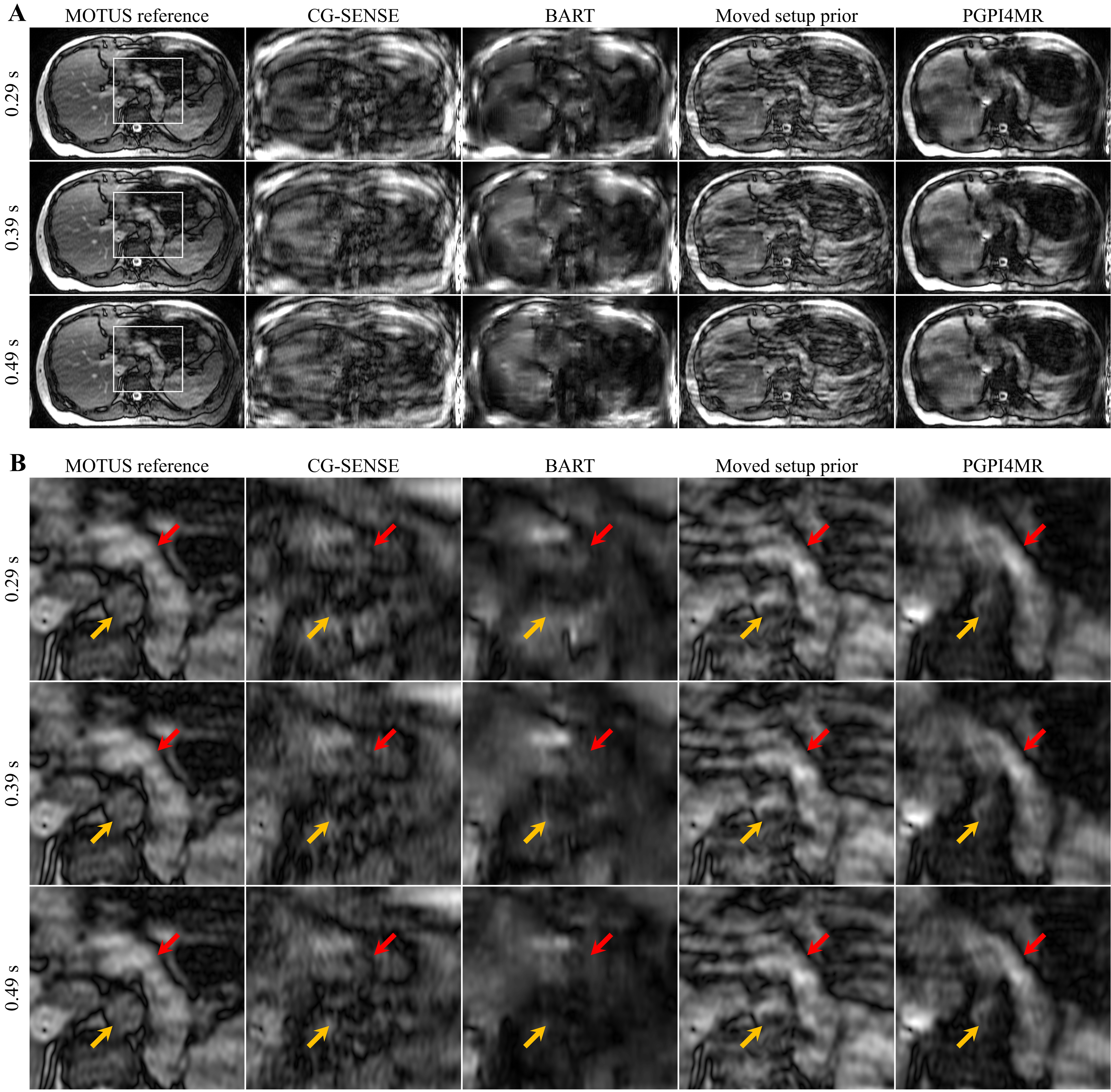}
\caption{Organs below the heart in sub-second 3D MRI of one volunteer (axial slice 93 mm below the liver dome). (A) In-vivo data, the first 0.29, 0.39, and 0.49 s windows after the setup scan (top to bottom); the box on the MOTUS reference marks the region enlarged in B. (B) The boxed region. Red arrows, the pancreas, identified visually on the MOTUS reference; amber arrows, the aorta, from the contour provided with the data. The arrows mark the same positions in every column of a row. The MOTUS reference is shown for anatomy only.}\label{fig:S1}
\end{figure}

\begin{figure}[p]
\centering
\includegraphics[width=0.75\textwidth,keepaspectratio]{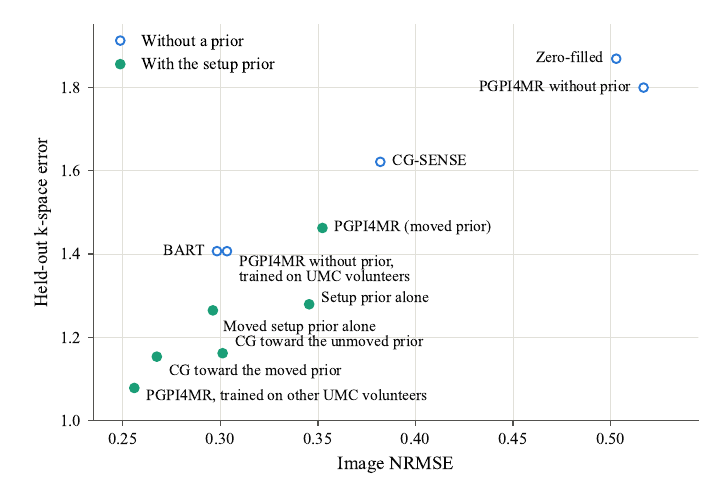}
\caption{Held-out k-space error against image NRMSE on the UMC thigh data (real k-space with fully sampled references; 0.69 s windows). Each point is the mean of one method over the 9 volunteers. Open circles, methods without a prior; filled circles, methods given the setup prior. Networks not labeled as trained on UMC volunteers were applied zero-shot. E2E-VarNet lies outside the plotted range (Table~\ref{tab:S4}).}\label{fig:S2}
\end{figure}

The MR-linac scans excited a slab of 200 mm (8 volunteers) or 220 mm (2 volunteers) centered on the heart, so the abdomen below the liver dome was covered only in part. In one volunteer, the slab reached the level of the pancreas. Fig.~\ref{fig:S1} shows an axial slice 93 mm below the liver dome in the first 0.29, 0.39, and 0.49 s windows after the setup scan, reconstructed by CG-SENSE, BART, and \methodname{}, which had not been trained on this volunteer, together with the moved setup prior that \methodname{} received. The aorta was identified from the contour provided with the data, and the pancreas visually on the MOTUS reference, because no pancreas contour is provided. \methodname{} and the moved setup prior alone showed both at every window length. In the CG-SENSE and BART images, undersampling artifacts obscured them. In these windows, \methodname{} had a held-out k-space error of 1.19, 1.18, and 1.22 at 0.49, 0.39, and 0.29 s. The corresponding errors were 1.82 to 1.92 for CG-SENSE, 1.67 to 1.80 for BART, 3.09 to 3.27 for E2E-VarNet, and 1.43 to 1.44 for the moved setup prior alone.

\clearpage

\clearpage
\begin{landscape}
\begin{table}[p]
\footnotesize
\caption{Per-volunteer results of every method. Arrows mark the better direction; bold, best; underlined, second best.}\label{tab:S1}
\setlength{\tabcolsep}{4pt}
\begin{tabular}{@{}p{2.5cm} l ccccc ccccc ccccc@{}}
\toprule
 & & \multicolumn{5}{c}{60 s} & \multicolumn{5}{c}{30 s} & \multicolumn{5}{c}{15 s} \\
\cmidrule(lr){3-7}\cmidrule(lr){8-12}\cmidrule(l){13-17}
Data & Volunteer & ZF & CG-S & BART & 3D & 4D & ZF & CG-S & BART & 3D & 4D & ZF & CG-S & BART & 3D & 4D \\
\midrule
\multirow{10}{=}{\raggedright 4D MRI, real data: held-out k-space error $\downarrow$} & 1 & 1.84 & 1.59 & \textbf{1.20} & 1.35 & \underline{1.24} & 1.90 & 1.70 & \underline{1.34} & 1.43 & \textbf{1.31} & 1.93 & 1.76 & \underline{1.50} & 1.52 & \textbf{1.45} \\
 & 2 & 1.83 & 1.61 & \underline{1.34} & 1.40 & \textbf{1.26} & 1.88 & 1.68 & \underline{1.46} & 1.48 & \textbf{1.35} & 1.94 & 1.79 & 1.62 & \underline{1.60} & \textbf{1.53} \\
 & 3 & 1.85 & 1.60 & \underline{1.28} & 1.37 & \textbf{1.27} & 1.92 & 1.75 & \underline{1.48} & 1.52 & \textbf{1.37} & 1.97 & 1.88 & \underline{1.65} & 1.66 & \textbf{1.50} \\
 & 4 & 1.84 & 1.64 & \underline{1.33} & 1.40 & \textbf{1.26} & 1.92 & 1.77 & \underline{1.48} & 1.50 & \textbf{1.33} & 1.95 & 1.82 & \underline{1.60} & 1.62 & \textbf{1.50} \\
 & 5 & 1.84 & 1.60 & \textbf{1.27} & 1.38 & \textbf{1.27} & 1.91 & 1.71 & \underline{1.48} & 1.53 & \textbf{1.41} & 1.95 & 1.80 & \underline{1.61} & \underline{1.61} & \textbf{1.53} \\
 & 6 & 1.88 & 1.61 & \textbf{1.23} & 1.37 & \underline{1.26} & 1.93 & 1.74 & \underline{1.42} & 1.51 & \textbf{1.36} & 1.97 & 1.84 & \underline{1.64} & 1.69 & \textbf{1.51} \\
 & 7 & 1.82 & 1.59 & \underline{1.27} & 1.37 & \textbf{1.23} & 1.88 & 1.72 & \underline{1.47} & 1.49 & \textbf{1.35} & 1.93 & 1.82 & \underline{1.63} & 1.66 & \textbf{1.51} \\
 & 8 & 1.86 & 1.68 & \underline{1.43} & 1.52 & \textbf{1.39} & 1.91 & 1.77 & \underline{1.55} & 1.64 & \textbf{1.46} & 1.95 & 1.85 & \underline{1.69} & 1.77 & \textbf{1.62} \\
 & 9 & 1.87 & 1.73 & \underline{1.54} & 1.63 & \textbf{1.48} & 1.93 & 1.82 & \underline{1.67} & 1.75 & \textbf{1.56} & 1.96 & 1.87 & \underline{1.76} & 1.86 & \textbf{1.69} \\
 & 10 & 1.89 & 1.67 & \underline{1.41} & 1.48 & \textbf{1.31} & 1.93 & 1.77 & \underline{1.56} & 1.62 & \textbf{1.43} & 1.96 & 1.84 & \underline{1.67} & 1.74 & \textbf{1.57} \\
\midrule
 & & \multicolumn{5}{c}{61 s} & \multicolumn{5}{c}{30 s} & \multicolumn{5}{c}{15 s} \\
\cmidrule(lr){3-7}\cmidrule(lr){8-12}\cmidrule(l){13-17}
 & Volunteer & ZF & CG-S & BART & 3D & 4D & ZF & CG-S & BART & 3D & 4D & ZF & CG-S & BART & 3D & 4D \\
\midrule
\multirow{10}{=}{\raggedright 4D MRI, simulation: NRMSE $\downarrow$} & 1 & 0.781 & 0.488 & \underline{0.270} & 0.289 & \textbf{0.219} & 0.867 & 0.658 & 0.440 & \underline{0.407} & \textbf{0.282} & 0.915 & 0.781 & 0.633 & \underline{0.568} & \textbf{0.403} \\
 & 2 & 0.778 & 0.495 & \underline{0.273} & 0.286 & \textbf{0.224} & 0.879 & 0.674 & 0.465 & \underline{0.418} & \textbf{0.296} & 0.926 & 0.791 & 0.654 & \underline{0.594} & \textbf{0.421} \\
 & 3 & 0.757 & 0.402 & \underline{0.176} & 0.190 & \textbf{0.137} & 0.870 & 0.626 & 0.370 & \underline{0.320} & \textbf{0.204} & 0.910 & 0.734 & 0.554 & \underline{0.485} & \textbf{0.303} \\
 & 4 & 0.780 & 0.481 & \underline{0.240} & 0.265 & \textbf{0.208} & 0.877 & 0.673 & 0.423 & \underline{0.402} & \textbf{0.285} & 0.926 & 0.788 & 0.623 & \underline{0.596} & \textbf{0.418} \\
 & 5 & 0.764 & 0.412 & \underline{0.185} & 0.203 & \textbf{0.149} & 0.882 & 0.647 & 0.377 & \underline{0.344} & \textbf{0.210} & 0.925 & 0.769 & 0.587 & \underline{0.535} & \textbf{0.317} \\
 & 6 & 0.771 & 0.440 & \underline{0.210} & 0.255 & \textbf{0.168} & 0.850 & 0.607 & \underline{0.350} & 0.382 & \textbf{0.230} & 0.906 & 0.747 & 0.565 & \underline{0.550} & \textbf{0.352} \\
 & 7 & 0.754 & 0.381 & \underline{0.154} & 0.187 & \textbf{0.123} & 0.854 & 0.590 & 0.340 & \underline{0.299} & \textbf{0.173} & 0.902 & 0.703 & 0.533 & \underline{0.464} & \textbf{0.299} \\
 & 8 & 0.758 & 0.515 & \underline{0.277} & 0.321 & \textbf{0.236} & 0.832 & 0.672 & \underline{0.451} & 0.483 & \textbf{0.322} & 0.882 & 0.772 & \underline{0.637} & 0.654 & \textbf{0.446} \\
 & 9 & 0.797 & 0.550 & \underline{0.307} & 0.384 & \textbf{0.247} & 0.890 & 0.722 & \underline{0.561} & 0.590 & \textbf{0.372} & 0.956 & 0.871 & \underline{0.801} & 0.809 & \textbf{0.611} \\
 & 10 & 0.781 & 0.461 & \underline{0.219} & 0.264 & \textbf{0.178} & 0.882 & 0.664 & \underline{0.429} & 0.433 & \textbf{0.254} & 0.932 & 0.781 & 0.629 & \underline{0.621} & \textbf{0.398} \\
\bottomrule
\end{tabular}
\tabnote{ZF, zero-filled; CG-S, CG-SENSE; 3D, \methodname{} without coupling (3D); 4D, \methodname{} (4D, phases coupled). Marks rank the methods of each volunteer at each scan time. Each volunteer was scored by a network that was not trained on it. Simulation: Poisson-disk sampling at the densities of 61, 30, and 15 s.}
\end{table}
\end{landscape}
\clearpage
\begin{landscape}
\begin{table}[p]
\footnotesize
\caption*{\textbf{Table S1.} Per-volunteer results of every method (continued)}
\setlength{\tabcolsep}{4pt}
\begin{tabular}{@{}p{3.6cm} l ccc@{}}
\toprule
 & & \multicolumn{3}{c}{MODEST, 4D MRI} \\
\cmidrule(l){3-5}
Data & Volunteer & 60 s (simulation: 61 s) & 30 s & 15 s \\
\midrule
\multirow{3}{=}{\raggedright Real data: held-out k-space error $\downarrow$} & 6 & 1.86 & 1.85 & 1.88 \\
 & 9 & 1.94 & 1.93 & 1.80 \\
 & 10 & 1.90 & 1.89 & 1.81 \\
\cdashline{1-5}
\multirow{3}{=}{\raggedright Simulation: NRMSE $\downarrow$} & \rule{0pt}{2.5ex}6 & 0.234 & 0.287 & 0.471 \\
 & 9 & 0.300 & 0.379 & 0.608 \\
 & 10 & 0.238 & 0.314 & 0.469 \\
\bottomrule
\end{tabular}
\tabnote{MODEST was trained on 5 volunteers and selected on 2 others, so it is shown only for the remaining 3 volunteers and was not ranked. Its simulated inputs were those of the other methods. It was trained at the 61 s sampling density only.}

\bigskip
\begin{tabular}{@{}p{1.9cm} l ccccccc ccccccc@{}}
\toprule
 & & \multicolumn{7}{c}{Real data: held-out k-space error $\downarrow$} & \multicolumn{7}{c}{Simulation: NRMSE $\downarrow$} \\
\cmidrule(lr){3-9}\cmidrule(l){10-16}
Task & Volunteer & ZF & CG-S & BART & E2E$^{\mathrm{a}}$ & NP & CGP & \methodname{} & ZF & CG-S & BART & E2E$^{\mathrm{b}}$ & NP & CGP & \methodname{} \\
\midrule
\multirow{10}{=}{\raggedright Sub-second MRI, 0.49 s windows} & 1 & 1.94 & 1.73 & 1.42 & 2.72 & 1.62 & \underline{1.21} & \textbf{0.95} & 0.921 & 0.807 & 0.708 & 0.870 & 0.754 & \underline{0.307} & \textbf{0.270} \\
 & 2 & 1.93 & 1.70 & 1.43 & 3.41 & 1.52 & \underline{1.22} & \textbf{1.04} & 0.903 & 0.756 & 0.636 & 0.855 & 0.590 & \underline{0.264} & \textbf{0.243} \\
 & 3 & 1.98 & 1.91 & 1.74 & 3.62 & 1.83 & \underline{1.42} & \textbf{1.17} & 0.864 & 0.675 & 0.545 & 0.965 & 0.628 & \underline{0.240} & \textbf{0.212} \\
 & 4 & 1.94 & 1.80 & 1.65 & 2.85 & 1.64 & \underline{1.34} & \textbf{1.09} & 0.893 & 0.746 & 0.626 & 0.916 & 0.634 & \underline{0.301} & \textbf{0.246} \\
 & 5 & 1.97 & 1.83 & 1.62 & 3.43 & 1.72 & \underline{1.28} & \textbf{1.15} & 0.909 & 0.740 & 0.606 & 0.869 & 0.616 & \underline{0.245} & \textbf{0.228} \\
 & 6 & 1.96 & 1.81 & 1.52 & 2.88 & 1.69 & \underline{1.26} & \textbf{1.03} & 0.899 & 0.741 & 0.601 & 0.861 & 0.666 & \underline{0.274} & \textbf{0.239} \\
 & 7 & 1.95 & 1.82 & 1.67 & 4.96 & 1.96 & \underline{1.29} & \textbf{1.13} & 0.848 & 0.644 & 0.514 & 1.136 & 0.640 & \underline{0.217} & \textbf{0.202} \\
 & 8 & 1.96 & 1.84 & 1.63 & 2.99 & 1.93 & \underline{1.38} & \textbf{1.18} & 0.842 & 0.712 & 0.618 & 0.999 & 0.785 & \underline{0.309} & \textbf{0.286} \\
 & 9 & 1.93 & 1.78 & 1.66 & 2.83 & 1.76 & \underline{1.34} & \textbf{1.23} & 0.895 & 0.770 & 0.699 & 1.064 & 0.763 & \underline{0.335} & \textbf{0.324} \\
 & 10 & 1.98 & 1.89 & 1.72 & 3.42 & 1.71 & \underline{1.36} & \textbf{1.14} & 0.877 & 0.722 & 0.590 & 0.880 & 0.603 & \underline{0.266} & \textbf{0.227} \\
\bottomrule
\end{tabular}
\tabnote{E2E, E2E-VarNet; NP, \methodname{} without prior; CGP, CG toward the moved prior; \methodname{}, \methodname{} (moved prior) with the learned field. Marks rank the methods of each volunteer for each measure; E2E$^{\mathrm{b}}$ cells were not ranked. The held-out error of each volunteer was computed from the held-out samples of all its windows together, and its NRMSE was averaged over its windows; \methodname{} values average 2 training runs. $^{\mathrm{a}}$From its own predicted k-space of each coil. $^{\mathrm{b}}$Magnitude NRMSE on the separately simulated test set of Supplementary Methods, S6; the other simulation columns are complex NRMSE.}
\end{table}
\end{landscape}
\clearpage
\begin{table}[p]
\footnotesize
\caption*{\textbf{Table S1.} Results of sub-second MRI in physics-based simulation under 3 kinds of motion (continued). Values are means over 10 volunteers; bold, best; underlined, second best.}
\setlength{\tabcolsep}{4pt}
\begin{tabular}{@{}p{1.9cm} p{1.7cm} l ccc@{}}
\toprule
 & & & \multicolumn{3}{c}{Simulation} \\
\cmidrule(l){4-6}
Motion & Metric & Method & 0.49 s & 0.39 s & 0.29 s \\
\midrule
\multirow{10}{=}{\raggedright Normal breathing} & \multirow{7}{=}{\raggedright NRMSE $\downarrow$} & CG-SENSE & 0.731 & 0.766 & 0.799 \\
 &  & Unmoved setup prior alone & 0.320 & 0.319 & 0.313 \\
 &  & \methodname{}, unmoved prior & 0.261 & 0.267 & 0.284 \\
 &  & Prior moved along the registration-based field, alone & 0.310 & 0.309 & 0.308 \\
 &  & \methodname{}, prior with registration-based field & \underline{0.251} & \underline{0.258} & \underline{0.277} \\
 &  & Prior moved along the learned field, alone & 0.305 & 0.305 & 0.303 \\
 &  & \methodname{}, prior with learned field & \textbf{0.248} & \textbf{0.255} & \textbf{0.275} \\
\cdashline{2-6}
 & \multirow{3}{=}{\raggedright Position error (mm)~$\downarrow$} & \rule{0pt}{2.5ex}\methodname{}, unmoved prior & 1.79 & 1.95 & 1.78 \\
 &  & \methodname{}, prior with registration-based field & \underline{0.56} & \underline{0.71} & \underline{0.86} \\
 &  & \methodname{}, prior with learned field & \textbf{0.54} & \textbf{0.68} & \textbf{0.74} \\
\cdashline{1-6}
\multirow{9}{=}{\raggedright Random deformation plus drift} & \multirow{6}{=}{\raggedright NRMSE $\downarrow$} & \rule{0pt}{2.5ex}Unmoved setup prior alone & 0.527 & 0.529 & 0.534 \\
 &  & \methodname{}, unmoved prior & 0.431 & 0.436 & \underline{0.449} \\
 &  & Prior moved along the registration-based field, alone & 0.512 & 0.516 & 0.523 \\
 &  & \methodname{}, prior with registration-based field & \textbf{0.426} & \textbf{0.432} & \textbf{0.447} \\
 &  & Prior moved along the learned field, alone & 0.513 & 0.516 & 0.524 \\
 &  & \methodname{}, prior with learned field & \underline{0.428} & \underline{0.434} & \underline{0.449} \\
\cdashline{2-6}
 & \multirow{3}{=}{\raggedright Position error (mm)~$\downarrow$} & \rule{0pt}{2.5ex}\methodname{}, unmoved prior & 2.70 & 3.05 & 2.87 \\
 &  & \methodname{}, prior with registration-based field & \underline{1.34} & \underline{1.33} & \underline{1.54} \\
 &  & \methodname{}, prior with learned field & \textbf{1.29} & \textbf{1.26} & \textbf{1.51} \\
\cdashline{1-6}
\multirow{9}{=}{\raggedright Breathing beyond the setup range} & \multirow{6}{=}{\raggedright NRMSE $\downarrow$} & \rule{0pt}{2.5ex}Unmoved setup prior alone & 0.384 & 0.382 & 0.382 \\
 &  & \methodname{}, unmoved prior & 0.309 & 0.314 & 0.332 \\
 &  & Prior moved along the registration-based field, alone & 0.338 & 0.337 & 0.337 \\
 &  & \methodname{}, prior with registration-based field & \underline{0.272} & \underline{0.278} & \underline{0.296} \\
 &  & Prior moved along the learned field, alone & 0.332 & 0.331 & 0.332 \\
 &  & \methodname{}, prior with learned field & \textbf{0.267} & \textbf{0.272} & \textbf{0.291} \\
\cdashline{2-6}
 & \multirow{3}{=}{\raggedright Position error (mm)~$\downarrow$} & \rule{0pt}{2.5ex}\methodname{}, unmoved prior & 4.31 & 4.30 & 4.65 \\
 &  & \methodname{}, prior with registration-based field & \textbf{1.84} & \textbf{1.80} & \textbf{1.95} \\
 &  & \methodname{}, prior with learned field & \underline{2.07} & \underline{2.00} & \underline{2.08} \\
\bottomrule
\end{tabular}
\tabnote{Marks rank the methods of each column within each block. Complex NRMSE; each prior alone was scaled to the true image. Position error: 95th percentile over the body in the worst of 16 windows (Supplementary Methods, S8).}
\end{table}

\clearpage
\begin{landscape}
\begin{table}[p]
\footnotesize
\caption{Sensitivity analyses in vivo: the main comparisons with the fold as the unit and without the development volunteers. Each value is the percentage by which the first method lowered the held-out k-space error of the second (negative, a higher error), with its 95\% confidence interval.}\label{tab:S2}
\setlength{\tabcolsep}{3pt}
\begin{tabular}{@{}>{\raggedright\arraybackslash}p{1.7cm} >{\raggedright\arraybackslash}p{5.4cm} l r@{\hspace{9pt}}c r@{\hspace{9pt}}c r@{\hspace{9pt}}c@{}}
\toprule
 & & & \multicolumn{2}{c}{60 s} & \multicolumn{2}{c}{30 s} & \multicolumn{2}{c}{15 s} \\
\cmidrule(lr){4-5}\cmidrule(lr){6-7}\cmidrule(l){8-9}
Task & Comparison & Unit & \% lower & 95\% CI & \% lower & 95\% CI & \% lower & 95\% CI \\
\midrule
\multirow[t]{8}{=}{4D MRI} & \multirow[t]{3}{=}{\methodname{} (4D, phases coupled) vs \methodname{} without coupling (3D), primary end point} & Volunteers ($n$ = 10) & 9.1\rlap{$^{*}$} & 8.1 to 10.2 & 9.9\rlap{$^{*}$} & 8.9 to 10.8 & 7.8\rlap{$^{*}$} & 6.2 to 9.4 \\
 & & Folds ($n$ = 5) & 9.1\rlap{$^{*}$} & 7.8 to 10.4 & 9.9\rlap{$^{*}$} & 8.9 to 10.8 & 7.8\rlap{$^{*}$} & 6.4 to 9.3 \\
 & & Without the 2 development volunteers ($n$ = 8) & 9.0\rlap{$^{*}$} & 7.7 to 10.3 & 9.7\rlap{$^{*}$} & 8.6 to 10.9 & 7.6\rlap{$^{*}$} & 5.5 to 9.7 \\
\cdashline{2-9}
 & \multirow[t]{3}{=}{\methodname{} (4D, phases coupled) vs BART} & \rule{0pt}{2.5ex}Volunteers ($n$ = 10) & 2.5\rlap{$^{*}$} & 0.1 to 4.9 & 6.5\rlap{$^{*}$} & 5.0 to 8.0 & 5.9\rlap{$^{*}$} & 4.5 to 7.3 \\
 & & Folds ($n$ = 5) & 2.5 & $-$0.7 to 5.6 & 6.5\rlap{$^{*}$} & 4.3 to 8.6 & 5.9\rlap{$^{*}$} & 3.9 to 7.9 \\
 & & Without the 2 development volunteers ($n$ = 8) & 2.3 & $-$0.9 to 5.4 & 6.4\rlap{$^{*}$} & 4.4 to 8.3 & 6.0\rlap{$^{*}$} & 4.2 to 7.7 \\
\cdashline{2-9}
 & \rule{0pt}{2.5ex}\methodname{} (4D) with shuffled phases vs \methodname{} without coupling (3D) & Volunteers ($n$ = 10) & 7.6\rlap{$^{*}$} & 6.4 to 8.8 & 8.6\rlap{$^{*}$} & 7.3 to 9.8 & 6.8\rlap{$^{*}$} & 5.0 to 8.5 \\
\cdashline{2-9}
 & \rule{0pt}{2.5ex}\methodname{} without coupling (3D) vs BART & Volunteers ($n$ = 10) & $-$7.3\rlap{$^{*}$} & $-$9.2 to $-$5.4 & $-$3.7\rlap{$^{*}$} & $-$5.2 to $-$2.3 & $-$2.1\rlap{$^{*}$} & $-$3.6 to $-$0.6 \\
\midrule
 & & & \multicolumn{2}{c}{\methodname{} vs CG toward} & \multicolumn{2}{c}{\methodname{}, learned vs} & & \\
 & & & \multicolumn{2}{c}{the moved prior} & \multicolumn{2}{c}{registration-based field} & & \\
\cmidrule(lr){4-5}\cmidrule(lr){6-7}
\multirow[t]{3}{=}{Sub-second MRI} & \multirow[t]{3}{=}{\methodname{} (moved prior), 0.49 s: comparisons in the column headers} & Volunteers ($n$ = 10) & 15.5\rlap{$^{*}$} & 12.6 to 18.3 & 1.11\rlap{$^{*}$} & 0.83 to 1.38 & & \\
 & & Folds ($n$ = 5) & 15.5\rlap{$^{*}$} & 12.4 to 18.5 & 1.11\rlap{$^{*}$} & 0.81 to 1.40 & & \\
 & & Without the 4 development volunteers \mbox{($n$ = 6)} & 15.5\rlap{$^{*}$} & 11.9 to 19.0 & 1.17\rlap{$^{*}$} & 0.72 to 1.61 & & \\
\midrule
 & & & \multicolumn{6}{l}{60 s (4D) or 0.49 s (sub-second)} \\
\cmidrule(l){4-9}
\multirow[t]{3}{=}{Other checks} & \methodname{} (4D) vs a wider 3D network with more parameters (2 folds, post hoc) & Volunteers$^{\dagger}$ & 10.1\rlap{$^{*}$} & 6.3 to 13.6 & & & & \\
\cdashline{2-9}
 & \rule{0pt}{2.5ex}\methodname{} (4D), phases in order vs shuffled (samples and forward model permuted together) & Volunteers & 1.6\rlap{$^{*}$} & 0.9 to 2.3 & & & & \\
\cdashline{2-9}
 & \rule{0pt}{2.5ex}Setup prior alone, moved along the learned field vs misplaced by three-quarters of its breathing range & Volunteers & 3.7\rlap{$^{*}$} & 2.1 to 5.3 & & & & \\
\bottomrule
\end{tabular}
\tabnote{$^{*}$\textit{P}~$<$~0.05 (2-sided paired t-test; exact values in Table S5). $^{\dagger}$4 volunteers of 2 folds; post hoc. Fold analyses average the 2 test volunteers of each fold first. Settings were tuned in part on these development volunteers (2 for 4D MRI and 4 for sub-second MRI).}
\end{table}
\end{landscape}

\clearpage

\clearpage
\begin{landscape}
\footnotesize
\setlength{\tabcolsep}{4pt}\renewcommand{\arraystretch}{0.92}
\begin{longtable}{@{}>{\raggedright\arraybackslash}p{2.9cm} l cc cc c@{}}
\caption{Compute time and image latency. Values are the compute time per window or volume and the latency of the slowest window, with the statistic given in each header.}\label{tab:S3}\\
\toprule
 & & \multicolumn{2}{c}{\begin{tabular}{@{}c@{}}Compute, ms $\downarrow$:\\ median / 95th percentile\end{tabular}} & \multicolumn{2}{c}{\begin{tabular}{@{}c@{}}Latency of the slowest window, s $\downarrow$:\\ from the middle / from the start\end{tabular}} & \\
\cmidrule(lr){3-4}\cmidrule(lr){5-6}
Window & Method & Ada & Blackwell & Ada & Blackwell & Windows (volunteers) \\
\midrule
\endfirsthead
\caption*{\textbf{Table S3.} Compute time and image latency (continued)}\\
\toprule
Window & Method & Ada & Blackwell & Ada & Blackwell & Windows (volunteers) \\
\midrule
\endhead
\multirow{7}{=}{\raggedright MR-linac, 3 frames (0.29 s)} & Zero-filled & $<$1 / $<$1 & $<$1 / $<$1 & 0.150 / 0.297 & 0.149 / 0.297 & 1,015 (10) \\
 & CG-SENSE & 31 / 39 & 20 / 22 & 0.190 / 0.338 & 0.170 / 0.318 & 1,015 (10) \\
 & BART & 673 / 705 & 586 / 660 & 0.870 / 1.018 & 0.825 / 0.973 & 405 (4) \\
 & E2E-VarNet & 204 / 288 & 132 / 173 & 0.462 / 0.610 & 0.321 / 0.469 & 1,015 (10) \\
 & \methodname{} without prior & 145 / 169 & 109 / 130 & 0.318 / 0.465 & 0.277 / 0.425 & 1,015 (10) \\
 & CG toward the moved prior & 78 / 98 & 47 / 56 & 0.257 / 0.405 & 0.204 / 0.351 & 1,015 (10) \\
 & \methodname{} (moved prior) & 188 / 212 & 135 / 159 & 0.361 / 0.509 & 0.308 / 0.455 & 1,015 (10) \\
\cdashline{1-7}
\multirow{7}{=}{\raggedright MR-linac, 4 frames (0.39 s)} & \rule{0pt}{2.5ex}Zero-filled & $<$1 / $<$1 & $<$1 / $<$1 & 0.199 / 0.396 & 0.198 / 0.396 & 759 (10) \\
 & CG-SENSE & 32 / 39 & 20 / 23 & 0.239 / 0.436 & 0.220 / 0.417 & 759 (10) \\
 & BART & 680 / 709 & 590 / 660 & 0.920 / 1.117 & 0.871 / 1.068 & 304 (4) \\
 & E2E-VarNet & 204 / 289 & 132 / 173 & 0.503 / 0.700 & 0.370 / 0.567 & 759 (10) \\
 & \methodname{} without prior & 145 / 169 & 109 / 130 & 0.368 / 0.565 & 0.327 / 0.524 & 759 (10) \\
 & CG toward the moved prior & 80 / 100 & 49 / 56 & 0.303 / 0.500 & 0.254 / 0.450 & 759 (10) \\
 & \methodname{} (moved prior) & 188 / 213 & 136 / 159 & 0.414 / 0.611 & 0.419 / 0.616 & 759 (10) \\
\cdashline{1-7}
\multirow{7}{=}{\raggedright MR-linac, 5 frames (0.49 s)} & \rule{0pt}{2.5ex}Zero-filled & $<$1 / $<$1 & $<$1 / $<$1 & 0.248 / 0.495 & 0.247 / 0.494 & 608 (10) \\
 & CG-SENSE & 31 / 39 & 20 / 23 & 0.288 / 0.535 & 0.269 / 0.516 & 608 (10) \\
 & BART & 658 / 708 & 582 / 663 & 0.967 / 1.214 & 0.925 / 1.172 & 243 (4) \\
 & E2E-VarNet & 208 / 289 & 132 / 173 & 0.550 / 0.796 & 0.419 / 0.666 & 608 (10) \\
 & \methodname{} without prior & 145 / 170 & 109 / 130 & 0.417 / 0.663 & 0.376 / 0.623 & 608 (10) \\
 & CG toward the moved prior & 81 / 101 & 48 / 57 & 0.355 / 0.601 & 0.303 / 0.550 & 608 (10) \\
 & \methodname{} (moved prior) & 188 / 213 & 136 / 160 & 0.465 / 0.711 & 0.406 / 0.652 & 608 (10) \\
\midrule
 & & \multicolumn{2}{c}{Compute, s $\downarrow$} & & & \\
\cmidrule(lr){3-4}
 & & Ada & Blackwell & & & \\
\midrule
\multirow{7}{=}{\raggedright 4D MRI: one 16-phase volume from 60 s, every readout position; median over volunteers} & Zero-filled & 0.13 & 0.07 & & & \\
 & CG-SENSE & 4.7 & 2.6 & & & \\
 & BART & 25.5 & 17.5 & & & \\
 & \parbox[t]{3.6cm}{\raggedright MODEST, one slab of 48 readout positions\strut} & 27.7 & 21.4 & & & \\
 & \methodname{} without coupling (3D) & 10.5 & 5.5 & & & \\
 & \methodname{} (4D, phases coupled) & 13.0 & 7.5 & & & \\
\bottomrule
\end{longtable}
\tabnote{ BART ran on the GPU, as a separate program. Latency excludes data transfer, target localization, and beam response.}
\end{landscape}

\clearpage
\begin{table}[p]
\footnotesize
\caption{External data sets. Values are the held-out k-space error of each Stanford scan, correlations ($r$) with the scanner navigator, and mean\,$\pm$\,standard deviation over the UMC volunteers. Paired effects are the percentage by which the first method lowered the error of the second. Arrows mark the better direction; bold, best; underlined, second best.}\label{tab:S4}
\setlength{\tabcolsep}{4pt}
\begin{tabular}{@{}l *{6}{>{\centering\arraybackslash}p{1.1cm}} >{\centering\arraybackslash}p{1.55cm}@{}}
\toprule
 & \multicolumn{7}{c}{Stanford abdomen, 6 scans with normal breathing:} \\
 & \multicolumn{7}{c}{held-out k-space error $\downarrow$} \\
\cmidrule(l){2-8}
Scan & ZF & CG-S & BART & E2E & NP & CGP & \methodname{} \\
\midrule
101 & 1.92 & 1.35 & \textbf{1.15} & 2.56 & 1.67 & \underline{1.17} & 1.18 \\
102 & 1.93 & 1.68 & 1.45 & 2.44 & 1.92 & \underline{1.42} & \textbf{1.41} \\
104 & 1.95 & 1.75 & \textbf{1.45} & 2.59 & 1.85 & \underline{1.46} & \underline{1.46} \\
106 & 1.92 & 1.58 & \textbf{1.29} & 2.47 & 1.67 & \underline{1.30} & 1.31 \\
108 & 1.96 & 1.78 & \textbf{1.50} & 2.45 & 2.04 & \underline{1.51} & 1.53 \\
110 & 1.89 & 1.51 & 1.28 & 2.85 & 1.63 & \textbf{1.20} & \underline{1.21} \\
\cdashline{1-8}
\rule{0pt}{2.5ex}Mean & 1.93 & 1.61 & \textbf{1.35} & 2.56 & 1.80 & \textbf{1.35} & \textbf{1.35} \\
\bottomrule
\end{tabular}

\bigskip
\begin{tabular}{@{}l l r r r@{}}
\toprule
 & & \multicolumn{3}{c}{Stanford abdomen, all 10 scans} \\
\cmidrule(l){3-5}
 & & \multicolumn{2}{c}{Agreement with the navigator, $r$ $\uparrow$} & Warp on vs off, \\
\cmidrule(lr){3-4}
Breathing & Scan & Learned field & Registration-based field & \% lower error \\
\midrule
\multirow[t]{6}{*}{Normal} & 101 & 0.464 & 0.966 & 0.14 \\
 & 102 & 0.238 & 0.596 & 0.55 \\
 & 104 & 0.541 & 0.930 & 0.38 \\
 & 106 & 0.898 & 0.955 & 0.07 \\
 & 108 & 0.952 & 0.971 & 2.89 \\
 & 110 & 0.938 & 0.958 & 0.18 \\
\cdashline{1-5}
\rule{0pt}{2.5ex}\multirow[t]{2}{*}{Deep} & 105 & 0.499 & 0.834 & 0.12\rlap{$^{\mathrm{b}}$} \\
 & 107 & $-$0.266 & 0.478 & 0.17\rlap{$^{\mathrm{b}}$} \\
\cdashline{1-5}
\rule{0pt}{2.5ex}Rapid shallow & 109 & 0.966 & 0.956 & 0.12\rlap{$^{\mathrm{b}}$} \\
\cdashline{1-5}
\rule{0pt}{2.5ex}Triggered & 103 & $-$0.201 & $-$0.358 & $-$0.60\rlap{$^{\mathrm{b}}$} \\
\bottomrule
\end{tabular}

\bigskip
\begin{tabular}{@{}>{\raggedright\arraybackslash}p{7.8cm} r@{\hspace{9pt}}c r@{\hspace{9pt}}c@{}}
\toprule
 & \multicolumn{4}{c}{UMC thigh, 9 volunteers, 0.69 s windows} \\
\cmidrule(l){2-5}
Method & \multicolumn{2}{c}{Held-out k-space error $\downarrow$} & \multicolumn{2}{c}{Image NRMSE $\downarrow$} \\
\midrule
Zero-filled & \multicolumn{2}{c}{1.87\,$\pm$\,0.02} & \multicolumn{2}{c}{0.503\,$\pm$\,0.06} \\
CG-SENSE & \multicolumn{2}{c}{1.62\,$\pm$\,0.01} & \multicolumn{2}{c}{0.382\,$\pm$\,0.03} \\
BART & \multicolumn{2}{c}{1.41\,$\pm$\,0.02} & \multicolumn{2}{c}{0.298\,$\pm$\,0.02} \\
E2E-VarNet & \multicolumn{2}{c}{4.58\,$\pm$\,0.28} & \multicolumn{2}{c}{0.838\,$\pm$\,0.02} \\
\methodname{} without prior & \multicolumn{2}{c}{1.80\,$\pm$\,0.03} & \multicolumn{2}{c}{0.517\,$\pm$\,0.02} \\
CG toward the moved prior & \multicolumn{2}{c}{\underline{1.15}\,$\pm$\,0.04} & \multicolumn{2}{c}{\underline{0.267}\,$\pm$\,0.02} \\
\methodname{} (moved prior) & \multicolumn{2}{c}{1.46\,$\pm$\,0.17} & \multicolumn{2}{c}{0.352\,$\pm$\,0.04} \\
\methodname{}, trained on other UMC volunteers & \multicolumn{2}{c}{\textbf{1.08\,$\pm$\,0.06}} & \multicolumn{2}{c}{\textbf{0.256\,$\pm$\,0.03}} \\
\cdashline{1-5}
\rule{0pt}{2.5ex}\textit{Paired, first method vs second} & \textit{\% lower} & \textit{95\% CI} & \textit{\% lower} & \textit{95\% CI} \\
\methodname{} (moved prior) vs CG toward the moved prior & $-$26.1\rlap{$^{*}$} & $-$38.1 to $-$15.1 & $-$31.5\rlap{$^{*}$} & $-$40.7 to $-$22.9 \\
\methodname{}, trained on other UMC volunteers, vs CG toward the moved prior & 6.6\rlap{$^{*}$} & 3.8 to 9.3 & 4.4 & $-$0.4 to 9.0 \\
\methodname{} (moved prior) vs BART & $-$3.4 & $-$13.7 to 6.1 & $-$17.7\rlap{$^{*}$} & $-$28.5 to $-$7.9 \\
\methodname{} without prior vs BART & $-$27.9\rlap{$^{*}$} & $-$30.2 to $-$25.6 & $-$73.5\rlap{$^{*}$} & $-$79.6 to $-$67.5 \\
BART vs CG toward the moved prior & $-$22.0\rlap{$^{*}$} & $-$25.1 to $-$18.9 & $-$11.7\rlap{$^{*}$} & $-$16.3 to $-$7.3 \\
\methodname{} (moved prior) vs E2E-VarNet & 68.2\rlap{$^{*}$} & 64.8 to 71.2 & 58.2\rlap{$^{*}$} & 54.1 to 61.8 \\
\methodname{} without prior vs E2E-VarNet & 60.7\rlap{$^{*}$} & 58.9 to 62.4 & 38.3\rlap{$^{*}$} & 36.6 to 40.0 \\
\bottomrule
\end{tabular}
\tabnote{ZF, zero-filled; CG-S, CG-SENSE; E2E, E2E-VarNet; NP, \methodname{} without prior; CGP, CG toward the moved prior; \methodname{}, \methodname{} (moved prior). $^{*}$\textit{P}~$<$~0.05 (Table S5). Networks were not retrained unless stated. $^{\mathrm{b}}$Outside the registered test.}
\end{table}

\begin{table}[p]
\footnotesize
\caption*{\textbf{Table S4.} External data sets (continued)}
\setlength{\tabcolsep}{3.6pt}
\begin{tabular}{@{}>{\raggedright\arraybackslash}p{1.9cm} l c c r@{\hspace{9pt}}c@{}}
\toprule
 & & \multicolumn{4}{c}{UMC thigh, 9 volunteers, 0.69 s windows (exploratory)} \\
\cmidrule(l){3-6}
 & & CG toward the & \methodname{}, trained on & \multicolumn{2}{c}{Paired result} \\
\cmidrule(l){5-6}
Scan & Score & moved prior & other UMC volunteers & \% lower & 95\% CI \\
\midrule
\multirow[t]{5}{=}{Dynamic scan} & Held-out k-space error $\downarrow$ & 1.15\,$\pm$\,0.04 & \textbf{1.08\,$\pm$\,0.06} & 6.6\rlap{$^{*}$} & 3.8 to 9.3 \\
 & Image NRMSE, all windows $\downarrow$ & 0.267\,$\pm$\,0.02 & \textbf{0.256\,$\pm$\,0.03} & 4.4 & $-$0.4 to 9.0 \\
 & Image NRMSE, 80~mm~Hg $\downarrow$ & 0.281\,$\pm$\,0.03 & \textbf{0.269\,$\pm$\,0.03} & 4.4\rlap{$^{*}$} & 0.2 to 8.3 \\
 & Shading-insensitive NRMSE $\downarrow$ & 0.356\,$\pm$\,0.04 & \textbf{0.328\,$\pm$\,0.04} & 8.1\rlap{$^{*}$} & 4.0 to 12.0 \\
 & 3D SSIM$^{\mathrm{c}}$ $\uparrow$ & 0.724\,$\pm$\,0.05 & \textbf{0.787\,$\pm$\,0.05} & +0.062\rlap{$^{*}$} & 0.040 to 0.085 \\
\cdashline{1-6}
\multirow[t]{3}{=}{Hybrid scan, 80~mm~Hg plateau} & \rule{0pt}{2.5ex}Image NRMSE $\downarrow$ & 0.237\,$\pm$\,0.03 & \textbf{0.202\,$\pm$\,0.03} & 14.9\rlap{$^{*}$} & 10.6 to 19.0 \\
 & Shading-insensitive NRMSE $\downarrow$ & 0.350\,$\pm$\,0.05 & \textbf{0.304\,$\pm$\,0.06} & 13.8\rlap{$^{*}$} & 8.6 to 18.7 \\
 & 3D SSIM$^{\mathrm{c}}$ $\uparrow$ & 0.757\,$\pm$\,0.04 & \textbf{0.840\,$\pm$\,0.03} & +0.083\rlap{$^{*}$} & 0.063 to 0.103 \\
\bottomrule
\end{tabular}
\tabnote{Values are mean\,$\pm$\,standard deviation; bold, the better of the two methods. $^{*}$\textit{P}~$<$~0.05 (Table S5). Exploratory: \methodname{} trained on other UMC volunteers. Paired result: \% lower error of \methodname{}; $^{\mathrm{c}}$for 3D SSIM, the difference.}
\end{table}

\clearpage
{\footnotesize
\setlength{\tabcolsep}{4pt}
\begin{longtable}{@{}>{\raggedright\arraybackslash}p{6.9cm} l c c c@{}}
\caption{Exact \textit{P} values of the comparisons in Tables 1, 2, S2, and S4 and of the registered analyses.}\label{tab:S5}\\
\toprule
Comparison (first vs second) & Data & $n$ & Effect (95\% CI) & \textit{P} \\
\midrule
\endfirsthead
\toprule
Comparison (first vs second) & Data & $n$ & Effect (95\% CI) & \textit{P} \\
\midrule
\endhead
\multicolumn{5}{@{}l}{\textit{Table 1: the best value of a column against the second best}}\\
\methodname{} (4D) vs BART, 60 s & Real & 10 & 2.52 (0.06 to 4.91) & 0.045 \\
\methodname{} (4D) vs BART, 30 s & Real & 10 & 6.49 (4.95 to 8.01) & $6.5 \times 10^{-6}$ \\
\methodname{} (4D) vs BART, 15 s & Real & 10 & 5.92 (4.51 to 7.30) & $6.4 \times 10^{-6}$ \\
\methodname{} (4D) vs BART, 61 s & Simulated & 10 & 18.67 (16.86 to 20.44) & $5.3 \times 10^{-9}$ \\
\methodname{} (4D) vs 3D network, 30 s & Simulated & 10 & 35.95 (32.36 to 39.35) & $1.8 \times 10^{-8}$ \\
\methodname{} (4D) vs 3D network, 15 s & Simulated & 10 & 33.20 (29.56 to 36.65) & $3.4 \times 10^{-8}$ \\
\methodname{} (moved prior) vs CG toward the moved prior, 0.49 s & Real & 10 & 15.49 (12.56 to 18.32) & $1.4 \times 10^{-6}$ \\
Same, 0.39 s & Real & 10 & 14.14 (11.13 to 17.05) & $3.5 \times 10^{-6}$ \\
Same, 0.29 s & Real & 10 & 11.50 (7.97 to 14.89) & $5.8 \times 10^{-5}$ \\
\methodname{} (moved prior) vs CG toward the moved prior, 0.49 s & Simulated & 10 & 10.29 (7.07 to 13.39) & $6.5 \times 10^{-5}$ \\
Same, 0.39 s & Simulated & 10 & 9.09 (5.36 to 12.66) & $4.5 \times 10^{-4}$ \\
Same, 0.29 s & Simulated & 10 & 3.40 ($-$3.36 to 9.71) & 0.28 \\
UMC: \methodname{}, trained, vs CG toward the moved prior; held-out error & Real & 9 & 6.58 (3.76 to 9.30) & $7.4 \times 10^{-4}$ \\
UMC: same; image NRMSE & Real & 9 & 4.44 ($-$0.38 to 9.02) & 0.066 \\
Stanford: \methodname{} vs BART (equal rounded means) & Real & 6 & 0.50 ($-$3.07 to 3.94) & 0.73 \\
Stanford: \methodname{} vs CG toward the moved prior & Real & 6 & $-$0.24 ($-$1.02 to 0.54) & 0.46 \\
\addlinespace
\multicolumn{5}{@{}l}{\textit{Table 2: each design choice (Holm-adjusted values in parentheses)}}\\
4D vs 3D network, 60 s (simulation: 61 s) & Real & 10 & 9.13 (8.08 to 10.16) & $1.5 \times 10^{-8}$ \\
 & Simulated & 10 & 28.79 (24.90 to 32.47) & $1.6 \times 10^{-7}$ \\
4D vs 3D network, 30 s & Real & 10 & 9.87 (8.93 to 10.79) & $2.9 \times 10^{-9}$ \\
 & Simulated & 10 & 35.95 (32.36 to 39.35) & $1.8 \times 10^{-8}$ \\
4D vs 3D network, 15 s & Real & 10 & 7.84 (6.24 to 9.41) & $2.0 \times 10^{-6}$ \\
 & Simulated & 10 & 33.20 (29.56 to 36.65) & $3.4 \times 10^{-8}$ \\
4D MRI: 3D network vs BART, 60 s & Real & 10 & $-$7.27 ($-$9.15 to $-$5.43) & $7.4 \times 10^{-6}$ \\
 & Simulated & 10 & $-$14.21 ($-$19.60 to $-$9.05) & $1.1 \times 10^{-4}$ \\
Sub-second: \methodname{} without prior vs BART & Real & 10 & $-$8.21 ($-$13.15 to $-$3.48) & 0.0031 \\
 & Simulated & 10 & $-$8.64 ($-$16.48 to $-$1.32) & 0.025 \\
\methodname{} with vs without the unmoved prior & Real & 10 & 31.81 (28.73 to 34.75) & $1.1 \times 10^{-8}$ \\
 & Simulated & 10 & 60.43 (56.90 to 63.67) & $1.5 \times 10^{-9}$ \\
Prior alone, moved along the learned field vs unmoved & Real & 10 & 2.39 (1.81 to 2.96) & $6.6 \times 10^{-6}$ ($1.3 \times 10^{-5}$) \\
 & Simulated & 10 & 4.32 (2.52 to 6.09) & $4.5 \times 10^{-4}$ \\
Prior alone, learned vs registration-based field & Real & 10 & 1.82 (1.56 to 2.08) & $8.1 \times 10^{-8}$ ($2.4 \times 10^{-7}$) \\
 & Simulated & 10 & 1.44 (0.86 to 2.01) & $3.2 \times 10^{-4}$ \\
CG toward the prior, registration-based field vs unmoved & Real & 10 & 0.51 (0.10 to 0.92) & 0.020 \\
 & Simulated & 10 & 2.66 (1.19 to 4.10) & 0.0028 \\
\methodname{} along the learned field vs unmoved prior & Real & 10 & 3.06 (2.42 to 3.69) & $2.1 \times 10^{-6}$ \\
 & Simulated & 10 & 4.82 (3.02 to 6.58) & $2.1 \times 10^{-4}$ \\
\methodname{}, learned vs registration-based field & Real & 10 & 1.11 (0.83 to 1.38) & $7.3 \times 10^{-6}$ \\
 & Simulated & 10 & 1.17 (0.34 to 2.00) & 0.011 \\
\methodname{} along the learned field, warp on vs switched off & Real & 10 & 1.44 (0.66 to 2.21) & 0.0024 (0.0024) \\
 & Simulated & 10 & 3.05 (0.51 to 5.53) & 0.024 \\
\methodname{} vs CG toward the moved prior & Real & 10 & 15.49 (12.56 to 18.32) & $1.4 \times 10^{-6}$ \\
 & Simulated & 10 & 10.29 (7.07 to 13.39) & $6.5 \times 10^{-5}$ \\
\addlinespace
\multicolumn{5}{@{}l}{\textit{Table S2: sensitivity analyses}}\\
\methodname{} (4D) vs 3D network, 60 s; volunteers & Real & 10 & 9.13 (8.08 to 10.16) & $1.5 \times 10^{-8}$ \\
\methodname{} (4D) vs 3D network, 60 s; folds & Real & 5 & 9.13 (7.83 to 10.40) & $4.7 \times 10^{-5}$ \\
\methodname{} (4D) vs 3D network, 60 s; without the 2 development volunteers & Real & 8 & 8.99 (7.71 to 10.26) & $9.4 \times 10^{-7}$ \\
\methodname{} (4D) vs BART, 60 s; volunteers & Real & 10 & 2.52 (0.06 to 4.91) & 0.045 \\
\methodname{} (4D) vs BART, 60 s; folds & Real & 5 & 2.52 ($-$0.66 to 5.59) & 0.092 \\
\methodname{} (4D) vs BART, 60 s; without the 2 development volunteers & Real & 8 & 2.29 ($-$0.94 to 5.42) & 0.14 \\
\methodname{} (4D), phases shuffled, vs 3D network, 60 s; volunteers & Real & 10 & 7.63 (6.41 to 8.82) & $2.4 \times 10^{-7}$ \\
3D network vs BART, 60 s; volunteers & Real & 10 & $-$7.27 ($-$9.15 to $-$5.43) & $7.4 \times 10^{-6}$ \\
\methodname{} (4D) vs 3D network, 30 s; volunteers & Real & 10 & 9.87 (8.93 to 10.79) & $2.9 \times 10^{-9}$ \\
\methodname{} (4D) vs 3D network, 30 s; folds & Real & 5 & 9.87 (8.94 to 10.78) & $9.2 \times 10^{-6}$ \\
\methodname{} (4D) vs 3D network, 30 s; without the 2 development volunteers & Real & 8 & 9.75 (8.58 to 10.90) & $2.9 \times 10^{-7}$ \\
\methodname{} (4D) vs BART, 30 s; volunteers & Real & 10 & 6.49 (4.95 to 8.01) & $6.5 \times 10^{-6}$ \\
\methodname{} (4D) vs BART, 30 s; folds & Real & 5 & 6.49 (4.32 to 8.62) & 0.0013 \\
\methodname{} (4D) vs BART, 30 s; without the 2 development volunteers & Real & 8 & 6.38 (4.38 to 8.33) & $1.5 \times 10^{-4}$ \\
\methodname{} (4D), phases shuffled, vs 3D network, 30 s; volunteers & Real & 10 & 8.58 (7.33 to 9.80) & $1.1 \times 10^{-7}$ \\
3D network vs BART, 30 s; volunteers & Real & 10 & $-$3.74 ($-$5.18 to $-$2.32) & $2.0 \times 10^{-4}$ \\
\methodname{} (4D) vs 3D network, 15 s; volunteers & Real & 10 & 7.84 (6.24 to 9.41) & $2.0 \times 10^{-6}$ \\
\methodname{} (4D) vs 3D network, 15 s; folds & Real & 5 & 7.84 (6.39 to 9.26) & $1.3 \times 10^{-4}$ \\
\methodname{} (4D) vs 3D network, 15 s; without the 2 development volunteers & Real & 8 & 7.63 (5.55 to 9.67) & $6.6 \times 10^{-5}$ \\
\methodname{} (4D) vs BART, 15 s; volunteers & Real & 10 & 5.92 (4.51 to 7.30) & $6.4 \times 10^{-6}$ \\
\methodname{} (4D) vs BART, 15 s; folds & Real & 5 & 5.92 (3.91 to 7.89) & 0.0013 \\
\methodname{} (4D) vs BART, 15 s; without the 2 development volunteers & Real & 8 & 5.96 (4.21 to 7.68) & $1.0 \times 10^{-4}$ \\
\methodname{} (4D), phases shuffled, vs 3D network, 15 s; volunteers & Real & 10 & 6.76 (4.98 to 8.50) & $1.5 \times 10^{-5}$ \\
3D network vs BART, 15 s; volunteers & Real & 10 & $-$2.08 ($-$3.58 to $-$0.60) & 0.011 \\
\methodname{} vs CG toward the moved prior, 0.49 s; volunteers & Real & 10 & 15.49 (12.56 to 18.32) & $1.4 \times 10^{-6}$ \\
\methodname{}, learned vs registration-based field, 0.49 s; volunteers & Real & 10 & 1.11 (0.83 to 1.38) & $7.3 \times 10^{-6}$ \\
\methodname{} vs CG toward the moved prior, 0.49 s; folds & Real & 5 & 15.49 (12.36 to 18.51) & $2.1 \times 10^{-4}$ \\
\methodname{}, learned vs registration-based field, 0.49 s; folds & Real & 5 & 1.11 (0.81 to 1.40) & $5.0 \times 10^{-4}$ \\
\methodname{} vs CG toward the moved prior, 0.49 s; without the 4 development volunteers & Real & 6 & 15.52 (11.88 to 19.00) & $1.5 \times 10^{-4}$ \\
\methodname{}, learned vs registration-based field, 0.49 s; without the 4 development volunteers & Real & 6 & 1.17 (0.72 to 1.61) & 0.0011 \\
\methodname{} (4D) vs a wider 3D network with more parameters (2 folds, post hoc) & Real & 4 & 10.05 (6.35 to 13.61) & 0.0036 \\
\methodname{} (4D), phases in order vs shuffled (samples and forward model permuted together) & Real & 10 & 1.62 (0.93 to 2.31) & $5.1 \times 10^{-4}$ \\
Setup prior alone, moved along the learned field vs misplaced by three-quarters of its breathing range & Real & 10 & 3.72 (2.13 to 5.29) & $5.5 \times 10^{-4}$ \\
\addlinespace
\multicolumn{5}{@{}p{\textwidth}@{}}{\textit{Registered analyses (Supplementary Methods, S9; intervals as registered)}}\\
Primary end point: \methodname{} (4D) vs 3D network, 60 s & Real & 10 & 9.13 (8.08 to 10.16) & $1.5 \times 10^{-8}$ \\
\methodname{} (4D) from 30 s vs 3D network from 60 s & Real & 10 & 2.21 (0.52 to 3.88) & 0.016 \\
\methodname{} (4D) from 15 s vs 3D network from 30 s & Real & 10 & 0.38 ($-$1.22 to 1.95) & 0.60 \\
Confirmatory: CG toward the prior, registration-based field vs unmoved & Real & 10 & 0.51 (0.10 to 0.92) & 0.020 \\
Same, without the 2 volunteers seen before registration & Real & 8 & 0.43 ($-$0.03 to 0.89) & 0.062 \\
Learned displacement field (earlier design): \methodname{} warp on vs switched off & Real & 10 & $-$1.64 ($-$3.60 to 0.29) & 0.087 (1.0) \\
Learned amplitude (earlier design): \methodname{} warp on vs switched off & Real & 10 & 0.34 ($-$0.31 to 0.98) & 0.27 \\
Prior alone, learned vs registration-based field & Real & 10 & 1.82 (1.48 to 2.16) & $8.1 \times 10^{-8}$ ($2.4 \times 10^{-7}$) \\
Prior alone, learned field vs unmoved & Real & 10 & 2.39 (1.70 to 3.07) & $6.6 \times 10^{-6}$ ($1.3 \times 10^{-5}$) \\
\methodname{}, learned field, warp on vs switched off & Real & 10 & 1.44 (0.66 to 2.21) & 0.0024 \\
Stanford: prior alone, learned vs registration-based field & Real & 6 & 0.91 (0.53 to 1.29) & $6.5 \times 10^{-4}$ (0.0013) \\
Stanford: \methodname{} warp on vs switched off & Real & 6 & 0.71 ($-$0.45 to 1.85) & 0.18 \\
Simulation: field error, learned minus registration-based field & Simulated & 10 & +0.041 ($-$0.018 to 0.099) & 0.15 \\
\addlinespace
\multicolumn{5}{@{}l}{\textit{Table S4: external data sets}}\\
Stanford: \methodname{} warp on vs switched off (registered) & Real & 6 & 0.71 ($-$0.45 to 1.85) & 0.18 \\
UMC: zero-shot \methodname{} vs CG toward the moved prior, held-out error & Real & 9 & $-$26.09 ($-$38.15 to $-$15.09) & $3.8 \times 10^{-4}$ \\
UMC: zero-shot \methodname{} vs CG toward the moved prior, image NRMSE & Real & 9 & $-$31.52 ($-$40.71 to $-$22.94) & $1.4 \times 10^{-5}$ \\
UMC: zero-shot \methodname{} vs BART, held-out error & Real & 9 & $-$3.36 ($-$13.71 to 6.05) & 0.45 \\
UMC: zero-shot \methodname{} vs BART, image NRMSE & Real & 9 & $-$17.75 ($-$28.47 to $-$7.92) & 0.0025 \\
UMC: zero-shot \methodname{} without prior vs BART, held-out error & Real & 9 & $-$27.89 ($-$30.22 to $-$25.61) & $1.1 \times 10^{-9}$ \\
UMC: zero-shot \methodname{} without prior vs BART, image NRMSE & Real & 9 & $-$73.47 ($-$79.64 to $-$67.52) & $3.6 \times 10^{-10}$ \\
UMC: BART vs CG toward the moved prior, held-out error & Real & 9 & $-$22.00 ($-$25.13 to $-$18.94) & $9.1 \times 10^{-8}$ \\
UMC: BART vs CG toward the moved prior, image NRMSE & Real & 9 & $-$11.70 ($-$16.32 to $-$7.26) & $2.3 \times 10^{-4}$ \\
UMC: zero-shot \methodname{} vs E2E-VarNet, held-out error & Real & 9 & 68.20 (64.85 to 71.24) & $4.6 \times 10^{-9}$ \\
UMC: zero-shot \methodname{} vs E2E-VarNet, image NRMSE & Real & 9 & 58.15 (54.12 to 61.83) & $2.0 \times 10^{-8}$ \\
UMC: zero-shot \methodname{} without prior vs E2E-VarNet, held-out error & Real & 9 & 60.66 (58.87 to 62.37) & $3.7 \times 10^{-11}$ \\
UMC: zero-shot \methodname{} without prior vs E2E-VarNet, image NRMSE & Real & 9 & 38.35 (36.61 to 40.04) & $1.7 \times 10^{-10}$ \\
UMC, trained \methodname{} vs CG toward the moved prior: held-out error & Real & 9 & 6.58 (3.76 to 9.30) & $7.4 \times 10^{-4}$ \\
UMC, trained \methodname{} vs CG toward the moved prior: image NRMSE, all windows & Real & 9 & 4.44 ($-$0.38 to 9.02) & 0.066 \\
UMC, trained \methodname{} vs CG toward the moved prior: image NRMSE, 80 mm Hg & Real & 9 & 4.38 (0.25 to 8.35) & 0.040 \\
UMC, trained \methodname{} vs CG toward the moved prior: shading-insensitive NRMSE & Real & 9 & 8.08 (4.00 to 11.99) & 0.0021 \\
UMC, trained \methodname{} vs CG toward the moved prior: 3D SSIM & Real & 9 & +0.062 (0.040 to 0.085) & $2.0 \times 10^{-4}$ \\
UMC, trained \methodname{} vs CG toward the moved prior: plateau, image NRMSE & Real & 9 & 14.88 (10.56 to 19.00) & $6.9 \times 10^{-5}$ \\
UMC, trained \methodname{} vs CG toward the moved prior: plateau, shading-insensitive NRMSE & Real & 9 & 13.79 (8.60 to 18.69) & $3.8 \times 10^{-4}$ \\
UMC, trained \methodname{} vs CG toward the moved prior: plateau, 3D SSIM & Real & 9 & +0.083 (0.063 to 0.103) & $1.1 \times 10^{-5}$ \\
\bottomrule
\multicolumn{5}{@{}p{\textwidth}@{}}{\scriptsize 2-sided paired t-tests on the log ratios (unit: volunteer; Stanford: scan; folds: fold). Only the 2 end points fixed in advance (Supplementary Methods, S9) are confirmatory; values in parentheses are Holm-adjusted (Table 2) or as registered (Supplementary Methods, S9); all others are unadjusted. A positive effect means a lower error of the first method; for 3D SSIM it is the difference (higher is better).}\\
\end{longtable}
}
\clearpage

\end{document}